\documentclass[10pt,twocolumn,letterpaper]{article}
\usepackage{cvpr}              

\usepackage{graphicx}
\usepackage{amsmath}
\usepackage{amssymb}
\usepackage{booktabs}
\usepackage{multicol}
\usepackage[accsupp]{axessibility}

\usepackage{url}
\usepackage{bbm}
\usepackage{amsmath,amssymb}
\usepackage{amsthm}
\usepackage{bm}
\usepackage[ruled, lined, linesnumbered, commentsnumbered, longend]{algorithm2e}
\usepackage[dvipsnames]{xcolor}
\usepackage{graphicx}
\usepackage{overpic}
\usepackage{mathtools}

\usepackage{listings}
\usepackage{color}
\usepackage{verbatim}
\usepackage{multirow}
\usepackage{booktabs}
\usepackage{enumitem}
\usepackage{pifont}
\newcommand{\cmark}{\textcolor{ForestGreen}{\ding{51}}}

\newcommand{\xmark}{\textcolor{red}{\ding{55}}}

\newcommand{\valpha}{\bm{\alpha}}
\newcommand{\vbeta}{\bm{\beta}}
\newcommand{\mCh}{\vc}
\newcommand{\mPh}{\vp}
\newcommand{\mTh}{\vt}
\newcommand{\mUh}{\vu}
\newcommand{\mPb}{\bar{\mP}}
\newcommand{\mEt}{\tilde{\mE}}
\newcommand{\mTt}{\tilde{\mT}}
\newcommand{\mPt}{\tilde{\mP}}
\newcommand{\diagP}{\diag(\mPh)}
\newcommand{\vbetat}{\tilde{\vbeta}}
\newcommand{\vbt}{\tilde{\vb}}
\newcommand{\vta}{\vt^{(a)}}
\newcommand{\vttb}{\tilde{\vt}^{(b)}}
\newcommand{\bbone}{\mathbbm{1}}
\newcommand{\diag}{\operatorname{diag}}
\newcommand{\PiE}{\bm{\Pi}_{\mE}}
\newcommand{\dPl}{\nabla_{\mPh}\ell}
\newcommand{\ddPl}{\nabla_{\mPh}^2\ell}
\newcommand{\pz}{\phantom{0}}

\newcommand{\wbary}{0.1}

\newtheorem{theorem}{Theorem}
\newtheorem{lemma}[theorem]{Lemma}

\newenvironment{sproof}{
  \proof}{\endproof}

\usepackage{amsmath,amsfonts,bm}

\def\eqref#1{equation~\ref{#1}}
\def\Eqref#1{Equation~\ref{#1}}
\def\1{\bm{1}}

\def\va{{\bm{a}}}
\def\vb{{\bm{b}}}
\def\vc{{\bm{c}}}

\def\vp{{\bm{p}}}

\def\vt{{\bm{t}}}
\def\vu{{\bm{u}}}
\def\vv{{\bm{v}}}
\def\vw{{\bm{w}}}
\def\vx{{\bm{x}}}
\def\vy{{\bm{y}}}
\def\vz{{\bm{z}}}

\def\mA{{\bm{A}}}

\def\mC{{\bm{C}}}
\def\mD{{\bm{D}}}
\def\mE{{\bm{E}}}
\def\mF{{\bm{F}}}

\def\mI{{\bm{I}}}
\def\mJ{{\bm{J}}}
\def\mK{{\bm{K}}}

\def\mP{{\bm{P}}}

\def\mS{{\bm{S}}}
\def\mT{{\bm{T}}}
\def\mU{{\bm{U}}}

\DeclareMathAlphabet{\mathsfit}{\encodingdefault}{\sfdefault}{m}{sl}
\SetMathAlphabet{\mathsfit}{bold}{\encodingdefault}{\sfdefault}{bx}{n}

\usepackage[pagebackref,breaklinks,colorlinks]{hyperref}

\usepackage[capitalize]{cleveref}
\crefname{section}{Sec.}{Secs.}
\Crefname{section}{Section}{Sections}
\Crefname{table}{Table}{Tables}
\crefname{table}{Tab.}{Tabs.}

\def\cvprPaperID{10155} 
\def\confName{CVPR}
\def\confYear{2022}

\begin{document}

\title{A Unified Framework for Implicit Sinkhorn Differentiation}

\author{Marvin Eisenberger$^{*}$, Aysim Toker$^{*}$, Laura Leal-Taixé$^{*}$,  Florian Bernard$^{\dagger}$, Daniel Cremers$^{*}$\\[5pt]
Technical University of Munich$^{*}$, University of Bonn$^{\dagger}$
}
\twocolumn[{
\renewcommand\twocolumn[1][]{#1}
\vspace{-5em}
\maketitle
\thispagestyle{empty}
\begin{center}
    \begin{overpic}
    [width=1.\textwidth]{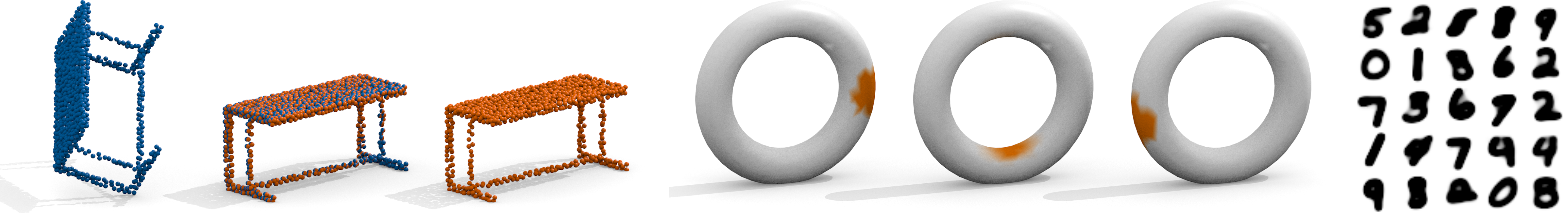}
    \put(9,-2){\small (a) Registration of two pointclouds}
    \put(43,-2){\small (b) Barycentric interpolation of distributions on a torus}
    \put(84,-2){\small (c) MNIST clustering}
    \end{overpic}
    \vspace{0.2em}
    \captionof{figure}{The Sinkhorn operator is becoming a fundamental building block for various computer vision algorithms. Relevant applications include (a) point cloud registration, (b) interpolation on manifolds, (c) image clustering, and many more \cite{sarlin2020superglue,yew2020rpm,yang2020mapping,liu2020learning,eisenberger2020deep}. A recent trend to training respective neural networks efficiently is implicit differentiation \cite{luise2018differential,flamary2018wasserstein,campbell2020solving,cuturi2020supervised,klatt2020empirical}. In this work, we provide a framework of implicit Sinkhorn differentiation that generalizes existing methods. It is the first to derive analytical gradients for the Sinkhorn operator in its most general form, covering all the applications (a)-(c) shown above.}
    \label{fig:teaser}
\end{center}
}]

\begin{abstract}

The Sinkhorn operator has recently experienced a surge of popularity in computer vision and related fields. One major reason is its ease of integration into deep learning frameworks. To allow for an efficient training of respective neural networks, we propose an algorithm that obtains analytical gradients of a Sinkhorn layer via implicit differentiation. In comparison to prior work, our framework is based on the most general formulation of the Sinkhorn operator. It allows for any type of loss function, while both the target capacities and cost matrices are differentiated jointly. We further construct error bounds of the resulting algorithm for approximate inputs. Finally, we demonstrate that for a number of applications, simply replacing automatic differentiation with our algorithm directly improves the stability and accuracy of the obtained gradients. Moreover, we show that it is computationally more efficient, particularly when resources like GPU memory are scarce.\footnote{Our implementation is available under the following link: \url{https://github.com/marvin-eisenberger/implicit-sinkhorn}}

\end{abstract}

\section{Introduction}

Computing matchings and permutations is a fundamental problem at the heart of many computer vision and machine learning algorithms. Common applications include pose estimation, 3D reconstruction, localization, information transfer, ranking, and sorting, with data domains ranging from images, voxel grids, point clouds, 3D surface meshes to generic Euclidean features. A popular tool to address this is the Sinkhorn operator, which has its roots in the theory of entropy regularized optimal transport~\cite{cuturi2013sinkhorn}. The Sinkhorn operator can be computed efficiently via a simple iterative matrix scaling approach. Furthermore, the resulting operator is differentiable, and can therefore be readily integrated into deep learning frameworks.

A key question is how to compute the first-order derivative of a respective Sinkhorn layer in practice.
The standard approach is automatic differentiation of Sinkhorn’s algorithm. Yet, this comes with a considerable computational burden because the runtime of the resulting backward pass scales linearly with the number of forward iterations. 
More importantly, since the computation graph needs to be maintained for all unrolled matrix-scaling steps, the memory demand is often prohibitively high for GPU processing.

A number of recent works leverage implicit gradients as an alternative to automatic differentiation~\cite{luise2018differential,flamary2018wasserstein,campbell2020solving,cuturi2020supervised,klatt2020empirical} to backpropagate through a Sinkhorn layer. Although such approaches prove to be computationally inexpensive, a downside is that corresponding algorithms are less straightforward to derive and implement.
Hence, many application works still rely on automatic differentiation~\cite{sarlin2020superglue,yew2020rpm,yang2020mapping,liu2020learning,eisenberger2020deep}.
Yet, the computational burden of automatic differentiation might drive practitioners to opt for an insufficiently small number of Sinkhorn iterations which in turn impairs the performance as we experimentally verify in \cref{sec:experiments}. 

To date, existing work on implicit differentiation of Sinkhorn layers suffers from two major limitations: (i) Most approaches derive gradients only for very specific settings, \ie specific loss functions, structured inputs, or only a subset of all inputs. Algorithms are therefore often not transferable to similar but distinct settings. (ii) Secondly, beyond their empirical success, there is a lack of an in-depth theoretical analysis that supports the use of implicit gradients.

\begin{figure*}
\vspace{-5pt}
\begin{center}
\includegraphics[width=0.9\linewidth]{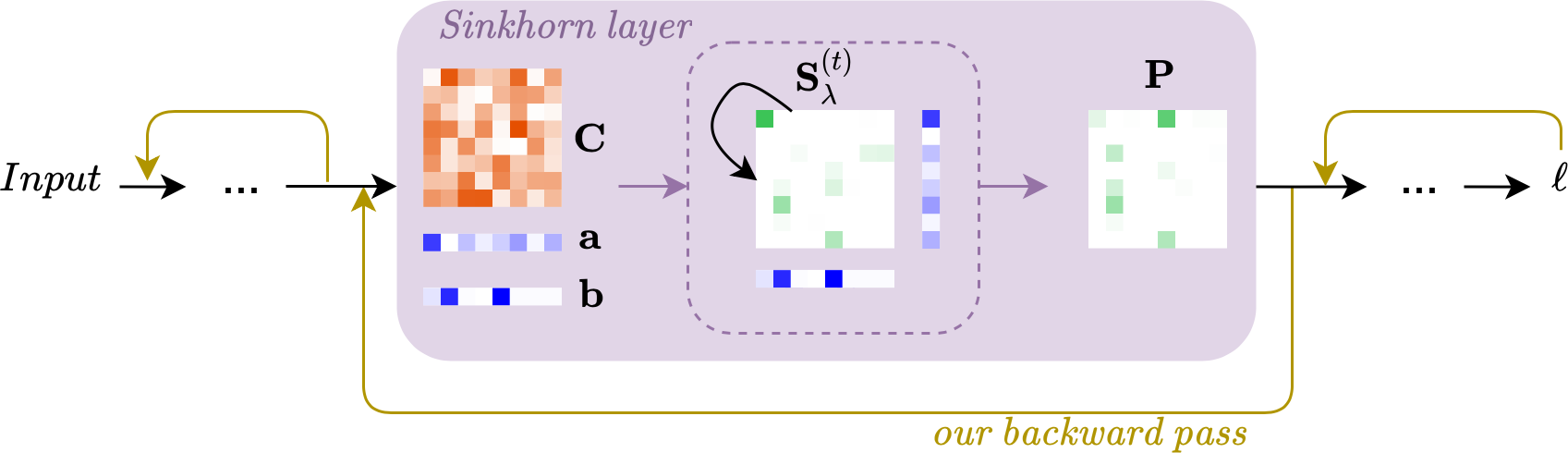}
\end{center}
\vspace{-5pt}
\caption{\textbf{Overview of a typical workflow with an embedded Sinkhorn layer.} We consider a neural network whose inputs are e.g. images, 3D point clouds, voxel grids, surface meshes, etc. The Sinkhorn layer maps the cost matrix $\mC$ and marginals $\va$, $\vb$ to the transportation plan $\mP$ via iterative matrix scaling. 
During training, we compute respective gradients $(\nabla_\mC\ell,\nabla_\va\ell,\nabla_\vb\ell)$ in closed form via implicit differentiation. Our algorithm applies to the most general formulation of the Sinkhorn operator: Both the cost matrix $\mC$ and marginals $\va,\vb$ are learnable and the whole network potentially contains learnable weights before and after the Sinkhorn layer.
}
\label{fig:overview}
\end{figure*}

Our work provides a unified framework of implicit differentiation techniques for Sinkhorn layers. To encourage practical adaptation, we provide a simple module that works out-of-the-box for the most general formulation, see \cref{fig:overview}. We can thus recover existing methods as special cases of our framework, see \cref{table:relatedsota} for an overview. Our contribution can be summarized as follows:
\begin{enumerate}
    \item From first principles we derive an efficient algorithm for computing gradients of a generic Sinkhorn layer. 
    \item We provide theoretical guarantees for the accuracy of the resulting gradients as a function of the approximation error in the forward pass (\cref{thm:errorbounds}). 
    \item Our PyTorch module can be applied in an out-of-the-box manner to existing approaches based on automatic differentiation. This often improves the quantitative results while using significantly less GPU memory.
\end{enumerate}

\section{Related work}\label{sec:relatedwork}

There is a vast literature on computational optimal transport (OT) \cite{villani2003topics,peyre2019computational}. In the following, we provide an overview of related machine learning applications. Our approach is based on entropy regularized optimal transport pioneered by \cite{cuturi2013sinkhorn}. 
The resulting differentiable Sinkhorn divergence can be used as a loss function for training machine learning models \cite{frogner2015learning,feydy2019interpolating,chizat2020faster}. To allow for first-order optimization, two common approaches for computing gradients are implicit differentiation \cite{luise2018differential,cuturi2020supervised,klatt2020empirical} and automatic differentiation \cite{genevay2018learning,ablin2020super}. 
Relevant applications of the Sinkhorn divergence include computing Wasserstein barycenters \cite{cuturi2014fast,solomon2015convolutional,luise2019sinkhorn}, dictionary learning \cite{schmitz2018wasserstein}, as well as using a geometrically meaningful loss function for autoencoders \cite{patrini2020sinkhorn} or generative adversarial networks (GAN) \cite{genevay2018learning,salimans2018improving}.

More recently, several approaches emerged that use the Sinkhorn operator as a differentiable transportation layer in a neural network. Potential applications include permutation learning \cite{santa2017deeppermnet,mena2018learning}, ranking \cite{adams2011ranking,cuturi2019differentiable}, sorting via reinforcement learning \cite{emami2018learning}, discriminant analysis \cite{flamary2018wasserstein} and computing matchings between images \cite{sarlin2020superglue}, point clouds \cite{yew2020rpm,yang2020mapping,liu2020learning} or triangle meshes \cite{eisenberger2020deep,pai2021fast}. Most of these approaches rely on automatic differentiation of the Sinkhorn algorithm to address the resulting bilevel optimization problem. In our work, we follow the recent trend of using implicit differentiation for the inner optimization layer \cite{amos2017optnet,gould2019deep,blondel2021efficient}. 
Other approaches compute the input cost matrix via Bayesian inverse modeling \cite{stuart2020inverse} or smooth the OT linear assignment problem (LAP) directly \cite{poganvcic2019differentiation}.

There are a number of methods that compute analytical gradients of a Sinkhorn layer, see \cref{table:relatedsota} for an overview. The idea of our work is to provide a unifying framework that generalizes specific methods, as well as providing additional theoretical insights. The pioneering work of Luise~\etal \cite{luise2018differential} computes gradients for the Sinkhorn divergence loss, while optimizing for the marginals. \cite{ablin2020super} and \cite{klatt2020empirical} provide further theoretical analysis. Flamary~\etal~\cite{flamary2018wasserstein} compute explicit gradients for the application of discriminant analysis. However, they directly solve the linear system specified by the implicit function theorem which leads to an algorithmic complexity of $\mathcal{O}(n^6)$. Similar to ours, \cite{campbell2020solving} and \cite{xie2020differentiable} compute gradients of the cost matrix $\mC$, but they assume constant marginals. The recent approach by Cuturi~\etal \cite{cuturi2020supervised} derives implicit gradients from the dual objective for the special case of low rank cost matrices $\mC(\vx,\vy)$.

\begin{table}
\begin{center}
\scalebox{0.80}{

\begin{tabular}{lccrcccc}
\toprule[0.2em]
Method & $\nabla_{[\va;\vb]}\ell$ & $\nabla_{\vx}\ell$ & $\nabla_{\mC}\ell$ & Loss & Obj. \\
\toprule[0.2em]
Luise~\etal~\cite{luise2018differential} & \cmark & \xmark & \xmark & Wasserstein & dual \\
Klatt~\etal~\cite{klatt2020empirical} & \cmark & \xmark & \xmark & Wasserstein & primal \\
Ablin~\etal~\cite{ablin2020super} & \cmark & \xmark & \xmark & Wasserstein & dual \\
Flamary~\etal~\cite{flamary2018wasserstein} & \xmark & \cmark & \xmark & Discr. analysis & primal \\
Campbell~\etal~\cite{campbell2020solving} & \xmark & \cmark & \cmark & any & primal \\
Xie~\etal~\cite{xie2020differentiable} & \xmark & \cmark & \cmark & any & dual \\
Cuturi~\etal~\cite{cuturi2020supervised} & \cmark & \cmark & \xmark & any & dual \\ \hline
\textbf{Ours} & \cmark & \cmark & \cmark & any & primal \\ \bottomrule[0.1em]
\end{tabular}
}
\centering
    \caption{\textbf{Overview of prior work.} We provide an overview of related approaches that, like ours, derive implicit gradients of a Sinkhorn layer. 
    For each method, we denote admissible inputs, \ie which inputs are differentiated. In the most general case, we want to optimize both the marginals $\va$ and $\vb$ and the cost matrix $\mC$ defined in \cref{sec:background}. As a special case, \cite{cuturi2020supervised,flamary2018wasserstein} provide gradients $\nabla_{\vx}\ell$ for low rank cost matrices of the form $\mC_{i,j}:=\|\vx_i-\vy_j\|_2^p$. We furthermore denote which types of loss functions are permitted and whether gradients are derived via the primal or dual objective.
    }
    \label{table:relatedsota}
\end{center}
\vspace{-5pt}
\end{table}

\section{Background}\label{sec:background}

\paragraph{Optimal transport.} Optimal transport enables us to compute the distance between two probability measures on the same domain $\Omega\subset\mathbb{R}^d$. In this work, we consider discrete probability measures $\mu:=\sum_{i=1}^ma_i\delta_{\vx_i}$ and $\nu:=\sum_{j=1}^nb_j\delta_{\vy_j}$, defined over the sets of points $\{\vx_1,\dots,\vx_m\}$ and $\{\vy_1,\dots,\vy_n\}$, where $\delta_{\vx_i}$ is the Dirac measure at $\vx_i$. Such measures are fully characterized by the probability mass vectors $\va\in\Delta_m$ and $\vb\in\Delta_n$ that lie on the probability simplex
\begin{equation}\label{eq:probsimplex}
    \Delta_m=\bigl\{\va\in\mathbb{R}^m|a_i\geq 0,\va^\top\bbone_m=1\bigr\},
\end{equation}
where $\bbone_m\in\mathbb{R}^m$ is the vector of all ones. We can then define the distance between $\mu$ and $\nu$ as
\begin{equation}\label{eq:otdistance}
    d(\mu,\nu):=\underset{\mP\in\Pi(\va,\vb)}{\min}\langle \mP,\mC\rangle_F.
\end{equation}
The transportation plan $\mP\in\Pi(\va,\vb)$ determines a discrete probability measure on the product space $\{\vx_1,\dots,\vx_m\}\times\{\vy_1,\dots,\vy_n\}$, whose marginal distributions coincide with $\mu$ and $\nu$. Consequently, $\mP$ is contained in the transportation polytope $\Pi(\va,\vb)$ defined as
\begin{equation}\label{eq:polytopepi}
    \Pi(\va,\vb):=\{\mP\in\mathbb{R}_+^{m\times n}|\mP\bbone_n=\va,\mP^\top\bbone_m=\vb\}.
\end{equation}
The cost matrix $\mC\in\mathbb{R}^{m\times n}$ specifies the transportation cost from individual points $\vx_i$ to $\vy_j$. Choosing $$\mC_{i,j}:=\|\vx_i-\vy_j\|_2^p$$ for $p\geq 1$, e.g. yields the so-called Wasserstein distance $d(\cdot,\cdot)=W^p_p(\cdot,\cdot)$, see \cite{villani2003topics}.

\paragraph{Entropy regularization.}
Evaluating the distance $d(\mu,\nu)$ in practice requires solving the linear assignment problem (LAP) from \cref{eq:otdistance}. This can be done via specialized algorithms like the Hungarian algorithm \cite{kuhn1955hungarian} or the Auction algorithm \cite{bertsekas1979distributed}, as well as recent solvers \cite{rubner1997earth,pele2009fast}. However, most approaches are computationally heavy and slow in practice~\cite{cuturi2013sinkhorn}. A popular alternative is augmenting the LAP objective in \cref{eq:otdistance} with an additional entropy regularizer, giving rise to the \emph{Sinkhorn operator}
\begin{equation}\label{eq:sinkhornoperator}
    S_\lambda(\mC,\va,\vb):=\underset{\mP\in\Pi(\va,\vb)}{\arg\min}\langle \mP,\mC\rangle_F-\lambda h(\mP),
\end{equation}
where $\lambda>0$ weights the regularization.
The seminal work of Cuturi~\etal \cite{cuturi2013sinkhorn} shows that the additional entropy regularization term $h(\mP)=-\sum_{i,j}P_{i,j}(\log P_{i,j}-1)$
allows for an efficient minimization of \cref{eq:sinkhornoperator}. Specifically, this can be done via a scheme of alternating Sinkhorn projections
\begin{align}\label{eq:sinkhornscheme}
    \mS^{(0)}_\lambda:=&\exp\biggl(-\frac{1}{\lambda} \mC\biggr)\nonumber,\quad\text{and}\\ 
    \mS^{(t+1)}_\lambda:=&\mathcal{T}_c\bigl(\mathcal{T}_r\bigl(\mS^{(t)}_\lambda\bigr)\bigr).
\end{align}
The operators $\mathcal{T}_c(\mS):=\mS\oslash (\bbone_m\bbone_m^\top \mS)\odot (\bbone_m\vb^\top)$ and $\mathcal{T}_r(\mS):=\mS\oslash (\mS\bbone_n\bbone_n^\top)\odot (\va\bbone_n^\top)$ correspond to renormalizations of the columns and rows of $\mS_\lambda^{(t)}$, where $\odot$ denotes the Hadamard product and $\oslash$ denotes element-wise division. As shown by \cite{cuturi2013sinkhorn}, in the limit this scheme converges to a minimizer $\mS^{(t)}_\lambda\xrightarrow{t\to\infty}\mS_\lambda$ of \cref{eq:sinkhornoperator}.
In practice, we can use a finite number of iterations $\tau\in\mathbb{N}$ to achieve a sufficiently small residual.

\section{Method} 

\subsection{Problem formulation}
Integrating the Sinkhorn operator from \cref{eq:sinkhornoperator} into deep neural networks has become a popular tool for a wide range of practical tasks, see our discussion in \cref{sec:relatedwork}. A major contributing factor is that the entropy regularization makes the mapping $S_\lambda:\mathbb{R}^{m\times n}\times\mathbb{R}^{m}\times\mathbb{R}^{n}\to\mathbb{R}^{m\times n}$ differentiable. To allow for first-order-optimization, we need to compute
\begin{align}
(\mC,\va,\vb)\qquad&\mapsto\qquad\mP^*:=S_\lambda(\mC,\va,\vb) \label{eq:forwardpass}\quad\text{and}\\
\nabla_\mP\ell\qquad&\mapsto\qquad(\nabla_\mC\ell,\nabla_\va\ell,\nabla_\vb\ell), \label{eq:backwardpass}
\end{align}
which denote the forward pass and the backpropagation of gradients, respectively.
Those expressions arise in the context of a typical workflow within a deep neural network  
with a scalar loss $\ell$ and learnable parameters before and/or after the Sinkhorn operator $S_\lambda$, see \cref{fig:overview} for an overview.

A common strategy is to replace the exact forward pass $S_\lambda(\mC,\va,\vb)$ in \cref{eq:forwardpass} by the approximate solution $\mS_\lambda^{(\tau)}$ from \cref{eq:sinkhornscheme}. 
Like the original solution in \cref{eq:sinkhornoperator}, $\mS_\lambda^{(\tau)}$ is differentiable w.r.t.~$(\mC,\va,\vb)$. Moreover, the mapping $(\mC,\va,\vb)\mapsto\mS_\lambda^{(\tau)}$ consists of a small number of matrix scaling operations that can be implemented in a few lines of code, see \cref{eq:sinkhornscheme}. 

\subsection{Backward pass via implicit differentiation}\label{subsec:backwardpassviaimplicitdifferentiation}
The goal of this section is to derive the main result stated in \cref{thm:closedformbackward}, which is the key motivation of our algorithm in \cref{subsec:algorithm}. To this end, we start by reframing the optimization problem in \cref{eq:sinkhornoperator} in terms of its Karush–Kuhn–Tucker (KKT) conditions, see \cref{subsec:prooflemmakkt} for a proof:
\begin{lemma}\label{thm:kkt}
The transportation plan $\mP^*$ is a global minimum of \cref{eq:sinkhornoperator} iff $\mathcal{K}(\mCh,\va,\vb,\mPh^*,\valpha^*,\vbeta^*)=\mathbf{0}_{l}$, with
\begin{equation}\label{eq:kkt}
    \mathcal{K}(\cdot):=
    \begin{bmatrix}
        \mCh+\lambda\log(\mPh^*)+\bbone_n\otimes\valpha^*+\vbeta^*\otimes\bbone_m\\
        (\bbone_n^\top\otimes\mI_m)\mPh^*-\va\\
        (\mI_n\otimes\bbone_m^\top)\mPh^*-\vb
    \end{bmatrix}
\end{equation}
where $l:=mn+m+n$. Here, $\valpha^*\in\mathbb{R}^m$ and $\vbeta^*\in\mathbb{R}^n$ are the dual variables corresponding to the two equality contraints in \cref{eq:polytopepi}.
We further define $\mCh,\mPh^*\in\mathbb{R}^{mn}$ as the vectorized versions of $\mC,\mP^*\in\mathbb{R}^{m\times n}$, respectively, and assume $\log(p):=-\infty,p\leq 0$.
\end{lemma}
Establishing this identity is an important first step towards computing a closed-form gradient for the backward pass in \cref{eq:backwardpass}. 
It reframes the optimization problem in \cref{eq:sinkhornoperator} as a root-finding problem $\mathcal{K}(\cdot)=\mathbf{0}$. In the next step, this then allows us to explicitly construct the derivative of the Sinkhorn operator $S_\lambda(\cdot)$ via implicit differentiation, see \cref{subsec:prooflemmaqp} for a proof:
\begin{lemma}\label{thm:qpjacobian}
The KKT conditions in \cref{eq:kkt} implicitly define a continuously differentiable function $(\mCh,\va,\vbt)\mapsto(\mPh,\valpha,\vbetat)$ with the Jacobian $\mJ\in\mathbb{R}^{(l-1)\times(l-1)}$ being
\begin{equation}\label{eq:qpjacobian}
    \mJ:=\frac
    {\partial\begin{bmatrix}
        \mPh;\valpha;\vbetat
    \end{bmatrix}}
    {\partial\begin{bmatrix}
        \mCh;-\va;-\vbt
    \end{bmatrix}}=-
    {\underbrace{\begin{bmatrix}
        \lambda \diag(\mPh)^{-1} & \mEt \\ \mEt^\top & \mathbf{0}
    \end{bmatrix}}_{:=\mK}}^{-1}.
\end{equation}
For brevity we use the short hand notation $[\vv;\vu]:=[\vv^\top,\vu^\top]^\top$ for stacking vectors $\vv,\vu$ vertically.
Note that the last entry of $\vbt:=\vb_{-n}$ and $\vbetat:=\vbeta_{-n}$ is removed. This is due to a surplus degree of freedom in the equality conditions from \cref{eq:polytopepi}, see part (b) of the proof. Likewise, for 
\begin{equation}
    \mE=\begin{bmatrix}\bbone_n\otimes\mI_m&\mI_n\otimes\bbone_m\end{bmatrix}\in\mathbb{R}^{mn\times(m+n)},
\end{equation}
the last column is removed $\mEt:=\mE_{:,-(m+n)}$.
\end{lemma}
In principle, we can use Lemma~\ref{thm:qpjacobian} directly to solve \cref{eq:backwardpass}. However, the computational cost of inverting the matrix $\mK$ in \cref{eq:qpjacobian} is prohibitive. In fact, even storing the Jacobian $\mJ$ in the working memory of a typical machine is problematic, since it is a dense matrix with $\mathcal{O}(mn)$ rows and columns, where $m,n>1000$ in practice. Instead, we observe that computing \cref{eq:backwardpass} merely requires us to compute vector-Jacobian products (VJP) of the form $\vv^\top\mJ$. The main results from this section can therefore be summarized as follows, see \cref{subsec:closedformbackward} for a proof:
\begin{theorem}[Backward pass]\label{thm:closedformbackward}
For $\mP=\mP^*$, the backward pass in \cref{eq:backwardpass} can be computed in closed form by solving the following linear system:
\begin{equation}\label{eq:closedformbackward}
    \begin{bmatrix}
        \lambda \diag(\mPh)^{-1} & \mEt \\ \mEt^\top & \mathbf{0}
    \end{bmatrix}
    \begin{bmatrix}
        \nabla_{\mCh}\ell\\-\nabla_{[\va;\vbt]}\ell
    \end{bmatrix}=
    \begin{bmatrix}
        -\nabla_{\mPh}\ell\\\mathbf{0}
    \end{bmatrix}.
\end{equation}
\end{theorem}

\subsection{Algorithm}\label{subsec:algorithm}

In the previous section, we derived a closed-form expression of the Sinkhorn backward pass in \cref{thm:closedformbackward}. This requires solving the sparse linear system in \cref{eq:closedformbackward}, which has $\mathcal{O}(mn)$ rows and columns, and thus amounts to a worst-case complexity of $\mathcal{O}(m^3n^3)$ \cite{flamary2018wasserstein}. We can further reduce the computation cost by exploiting the specific block structure of $\mK$, which leads to the following algorithm:

\begin{algorithm}
    \SetKwFunction{isOddNumber}{isOddNumber}
    \SetKwInOut{KwIn}{Input}
    \SetKwInOut{KwOut}{Output}
    \SetKwRepeat{Do}{do}{while}

    \KwIn{$\nabla_\mP\ell,\mP,\va,\vb$}
    \KwOut{$\nabla_\mC\ell,\nabla_\va\ell,\nabla_{\vb}\ell$}
    {

    $\mT\leftarrow\mP\odot\nabla_\mP\ell$.\label{alg:lnT}
    
    $\mTt\leftarrow\mT_{:,-n},\mPt\leftarrow\mP_{:,-n}\in\mathbb{R}^{m\times n-1}$.\label{alg:lnTt}
    
    $\vta\leftarrow\mT\bbone_n,\vttb\leftarrow\mTt^\top\bbone_m$.\label{alg:vtavttb}
    
    $\begin{bmatrix}\nabla_\va\ell\\\nabla_{\vbt}\ell\end{bmatrix}\leftarrow\begin{bmatrix}\diag(\va) & \mPt \\ \mPt^\top & \diag(\vbt) \end{bmatrix}^{-1}\begin{bmatrix}\vta\\\vttb\end{bmatrix}$.\label{alg:gradagradb}
    
    $\nabla_{\vb}\ell\leftarrow\begin{bmatrix}\nabla_{\vbt}\ell;0\end{bmatrix}$.\label{alg:gradbresidual}
    
    $\mU\leftarrow\nabla_\va\ell\bbone_n^\top+\bbone_m{\nabla_{\vb}\ell}^\top$.\label{alg:Umatrix}
    
    $\nabla_\mC\ell\leftarrow-\lambda^{-1}(\mT-\mP\odot\mU)$.\label{alg:gradm}
    }
    \caption{\emph{Sinkhorn operator backward}
    }\label{alg:backward}
\end{algorithm}

See \cref{sec:pytorchimplementation} for a PyTorch implementation of this algorithm. 
Most methods listed in \cref{table:relatedsota} consider a special case of the functional specified in \cref{eq:sinkhornoperator}. The resulting gradients of \cref{alg:backward} are thereby, for the most part, consistent with such specialized approaches.
We now show that the resulting gradients $\nabla_\mC\ell,\nabla_\va\ell,\nabla_\vb\ell$ from \cref{alg:backward} are indeed solutions of the linear system in \cref{thm:closedformbackward}.

\begin{theorem}\label{thm:algorithmequivalence}
Let $\va,\vb$ be two input marginals and $\mP=\mP^*$ the transportation plan resulting from the forward pass in \cref{eq:forwardpass}, then \cref{alg:backward} solves the backward pass \cref{eq:backwardpass}.
\end{theorem}
\begin{sproof} 
The main idea of this proof is showing that \cref{alg:backward} yields a solution $\nabla_{[\mCh;\va;\vbt]}\ell$ of the linear system from \cref{eq:closedformbackward}. To that end, we leverage the Schur complement trick which yields the following two expressions:
\begin{subequations}
\begin{equation}\label{eq:gradientla}
    \nabla_{[\va;\vbt]}\ell=\bigl(\mEt^\top\diagP\mEt\bigr)^{-1}\mEt^\top\diagP\nabla_{\mPh}\ell.
\end{equation}
\begin{equation}\label{eq:gradientlb}
    \nabla_{\mCh}\ell=
        -\lambda^{-1}\bigl(\diagP\nabla_{\mPh}\ell-\diagP\mEt\nabla_{[\va;\vbt]}\ell\bigr).
\end{equation}
\end{subequations}
In \cref{subsec:algorithmequivalence} we further show that these two identities in their vectorized form are equivalent to \cref{alg:backward}.
\end{sproof}

\subsection{Practical considerations}\label{subsec:practicalconsiderations}

\paragraph{Error bounds.} \cref{thm:algorithmequivalence} proves that \cref{alg:backward} computes the exact gradients $\nabla_\mC\ell,\nabla_\va\ell,\nabla_\vb\ell$, given that $\mP=\mP^*$ is the exact solution of \cref{eq:sinkhornoperator}. In practice, the operator $S_\lambda$ in \cref{eq:forwardpass} is replaced by the Sinkhorn approximation $\mS_\lambda^{(\tau)}$ from \cref{eq:sinkhornscheme} for a fixed, finite $\tau\in\mathbb{N}$. This small discrepancy in the approximation $\mP=\mS_\lambda^{(\tau)}\approx\mP^*$ propagates to the backward pass as follows:

\begin{theorem}[Error bounds]\label{thm:errorbounds}
Let $\mP^*:=S_\lambda(\mC,\va,\vb)$ be the exact solution of \cref{eq:sinkhornoperator} and let $\mP^{(\tau)}:=\mS_\lambda^{(\tau)}$ be the Sinkhorn estimate from \cref{eq:sinkhornscheme}. Further, let $\sigma_+,\sigma_-,C_1,C_2,\epsilon>0$, s.t. $\bigl\|\mP^*-\mP^{(\tau)}\bigr\|_F<\epsilon$ and that for all $\mP$ for which $\|\mP-\mP^*\|_F<\epsilon$ we have $\min_{i,j}\mP_{i,j}\geq\sigma_-$, $\max_{i,j}\mP_{i,j}\leq\sigma_+$ and the loss $\ell$ has bounded derivatives $\bigl\|\dPl\bigr\|_2\leq C_1$ and $\bigl\|\ddPl\bigr\|_F\leq C_2$. For $\kappa=\|\mEt^\dagger\|_2$, where $\mEt^\dagger$ indicates the Moore-Penrose inverse of $\mEt$, the difference between the gradients $\nabla_\mC\ell^*,\nabla_\va\ell^*,\nabla_\vb\ell^*$ of the exact $\mP^*$ and the gradients $\nabla_\mC\ell^{(\tau)},\nabla_\va\ell^{(\tau)},\nabla_\vb\ell^{(\tau)}$ of the approximate $\mP^{(\tau)}$, obtained via \cref{alg:backward}, satisfy
\begin{subequations}\label{eq:errorbounds}
\begin{equation}
\begin{multlined}\label{eq:errorboundsa}
    \bigl\|\nabla_{[\va;\vb]}\ell^*-\nabla_{[\va;\vb]}\ell^{(\tau)}\bigr\|_F\leq
    \\
    \kappa\sqrt{\frac{\sigma_+}{\sigma_-}}\biggl(\frac{1}{\sigma_-}C_1+C_2\biggr)\bigl\|\mP^*-\mP^{(\tau)}\bigr\|_F
\end{multlined}
\end{equation}
\begin{equation}
\begin{multlined}\label{eq:errorboundsb}
    \bigl\|\nabla_\mC\ell^*-\nabla_\mC\ell^{(\tau)}\bigr\|_F\leq
    \\
    \lambda^{-1}\sigma_+\biggl(\frac{1}{\sigma_-}C_1+C_2\biggr)\bigl\|\mP^*-\mP^{(\tau)}\bigr\|_F.
\end{multlined}
\end{equation}
\end{subequations}
\end{theorem}
We provide a proof in \cref{subsec:errorboundsproof}, as well as an empirical evaluation in \cref{subsec:gradientaccuracy}.

\paragraph{Computation cost.} In comparison to automatic differentiation (AD), the computation cost of \cref{alg:backward} is independent of the number of Sinkhorn iterations $\tau$. For square matrices, $m=n$, the runtime and memory complexities of AD are $\mathcal{O}(\tau n^2)$. On the other hand, our approach has a runtime and memory complexity of $\mathcal{O}(n^3)$ and $\mathcal{O}(n^2)$ respectively. We show empirical comparisons between the two approaches in \cref{subsec:computationalcomplexity}. 
Another compelling feature of our approach is that none of the operations in \cref{alg:backward} explicitly convert the matrices $\mP,\nabla_\mP\ell,\nabla_\mC\ell,\dots\in\mathbb{R}^{m\times n}$ into their vectorized form $\mPh,\nabla_\mPh\ell,\nabla_\mCh\ell,\dots\in\mathbb{R}^{mn}$.
This makes it computationally more efficient since GPU processing favors small, dense matrix operations over the large, sparse linear system in \cref{eq:closedformbackward}.

\paragraph{Marginal probability invariance.} As discussed in Lemma~\ref{thm:qpjacobian}, the last element of $\vbt$ needs to be removed to make $\mK$ invertible. However, setting the last entry of the gradient $\nabla_{b_n}\ell=0$ to zero still yields exact gradients: By definition, the full marginal $\vb$ is constrained to the probability simplex $\Delta_n$, see \cref{eq:probsimplex}.
In practice, we apply an a priori $\mathrm{softmax}$ to $\vb$ (and analogously $\va$). For some applications, $\vb$ can be assumed to be immutable, if we only want to learn the cost matrix $\mC$ and not the marginals $\va$ and $\vb$. Overall, this means that the gradient of $\vb$ is effectively indifferent to constant offsets of all entries, and setting $\nabla_{b_n}\ell=0$ does not contradict the statement of \cref{thm:closedformbackward}.

\begin{figure*}
\begin{center}
\includegraphics[width=0.31\textwidth]{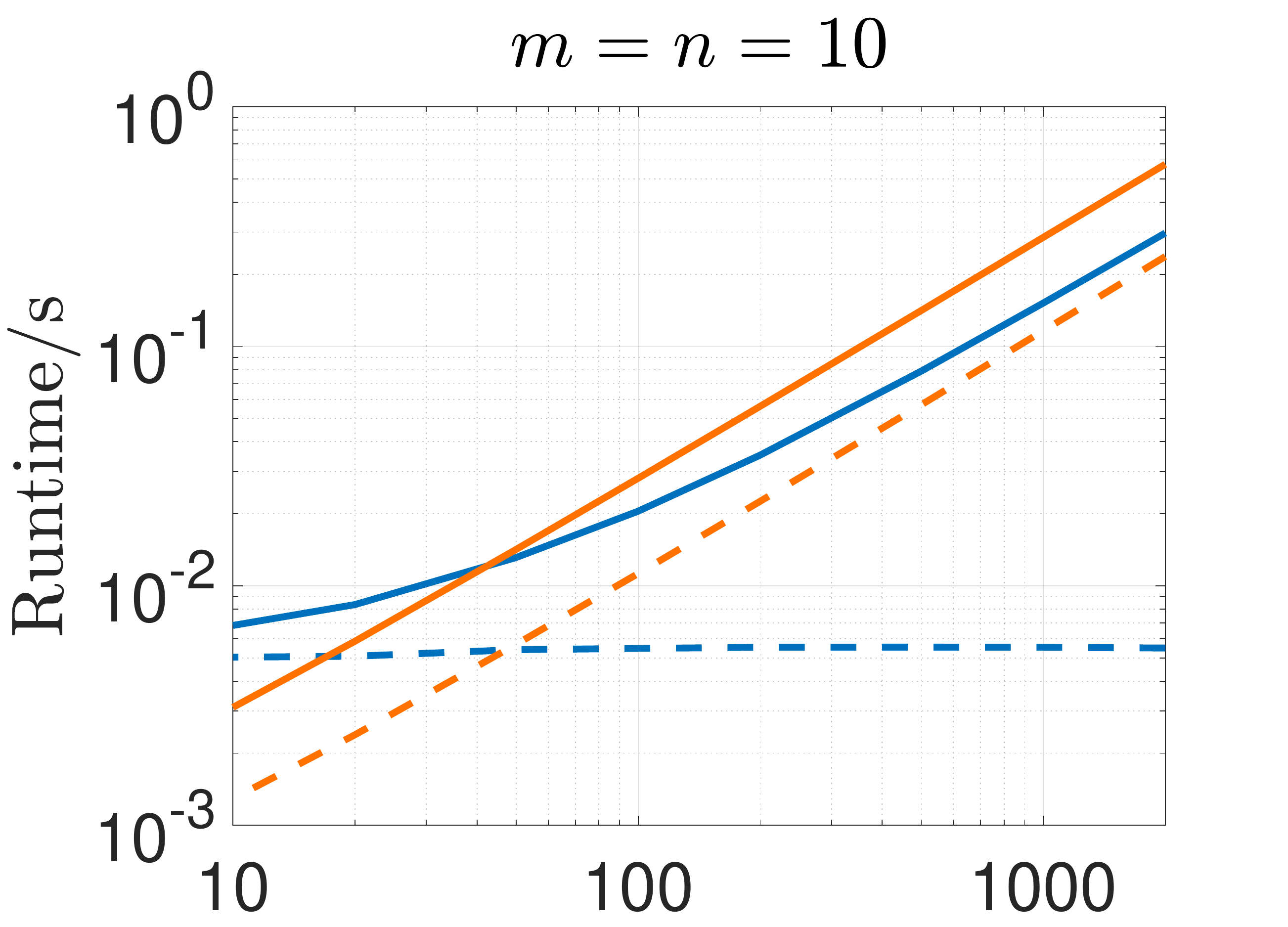}
\includegraphics[width=0.31\textwidth]{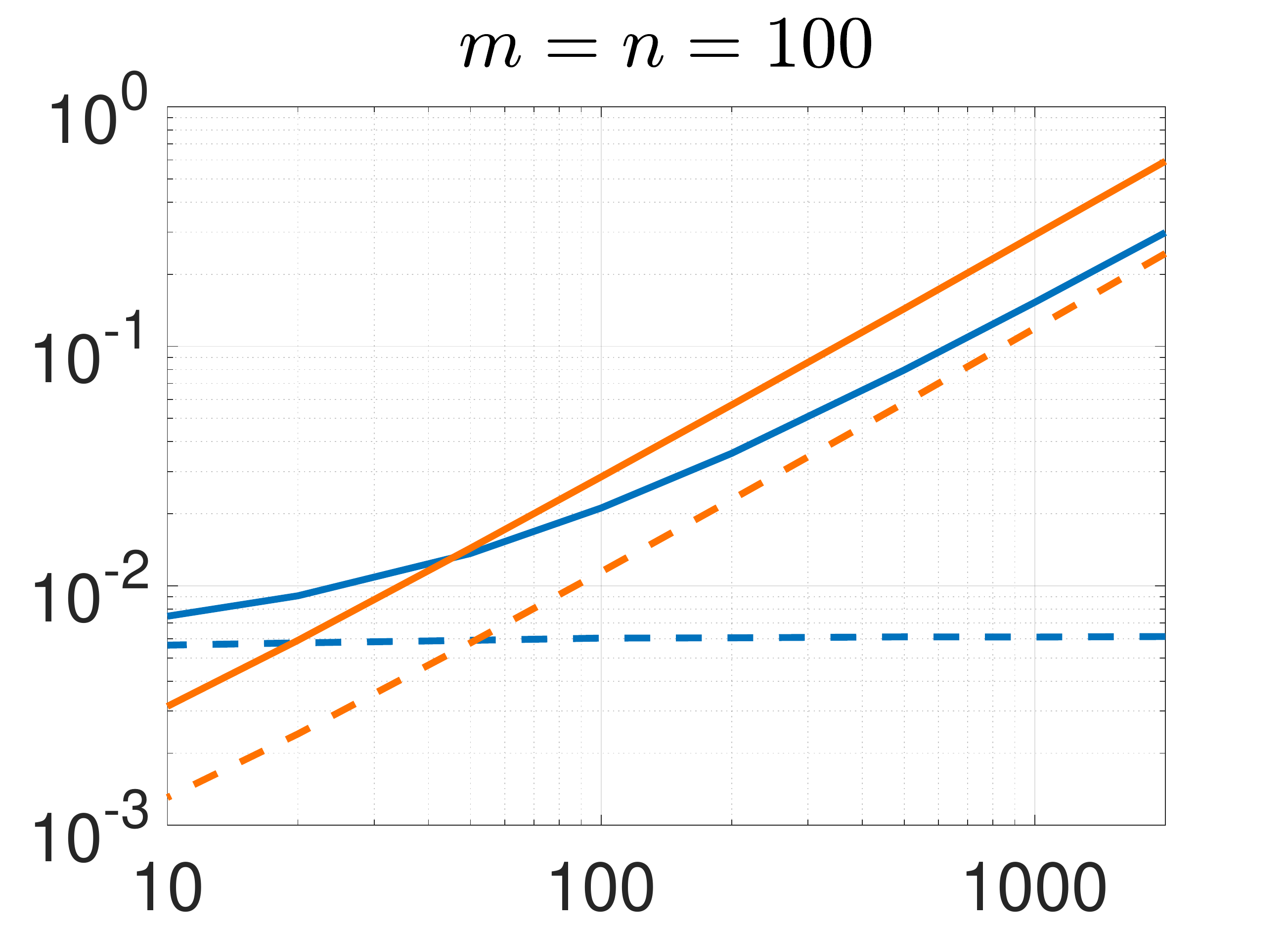}
\includegraphics[width=0.31\textwidth]{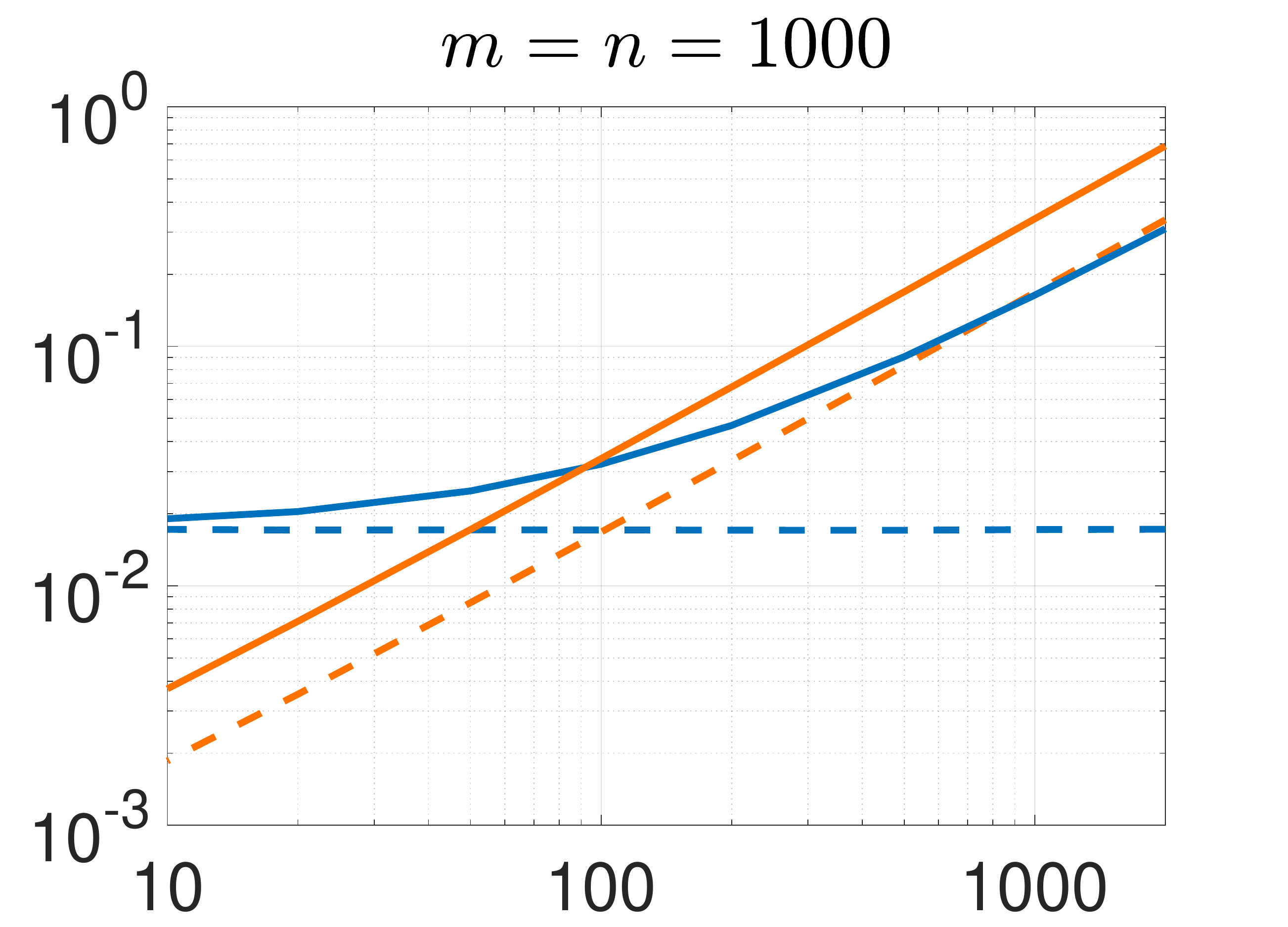}

\includegraphics[width=0.31\textwidth]{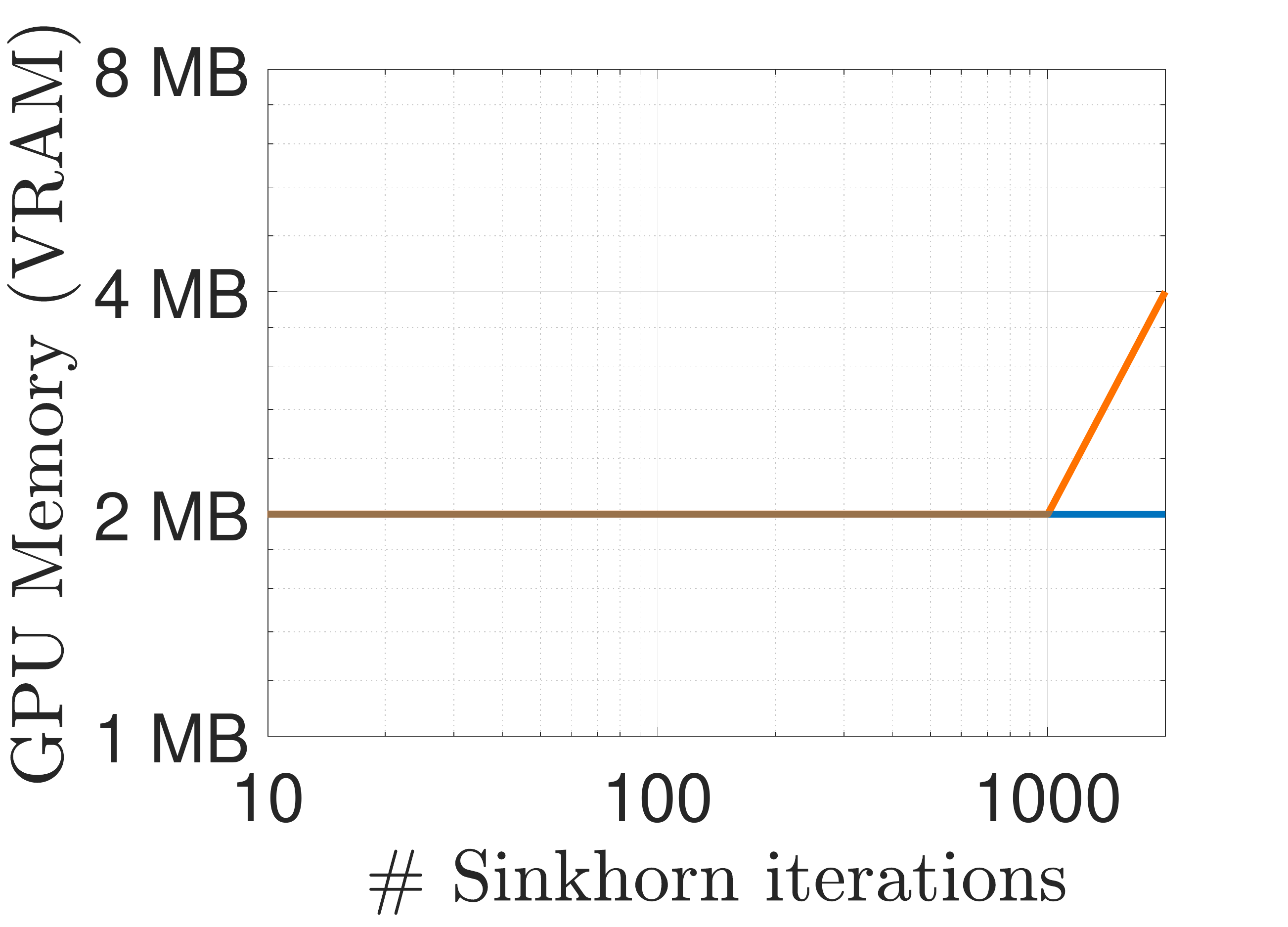}
\includegraphics[width=0.31\textwidth]{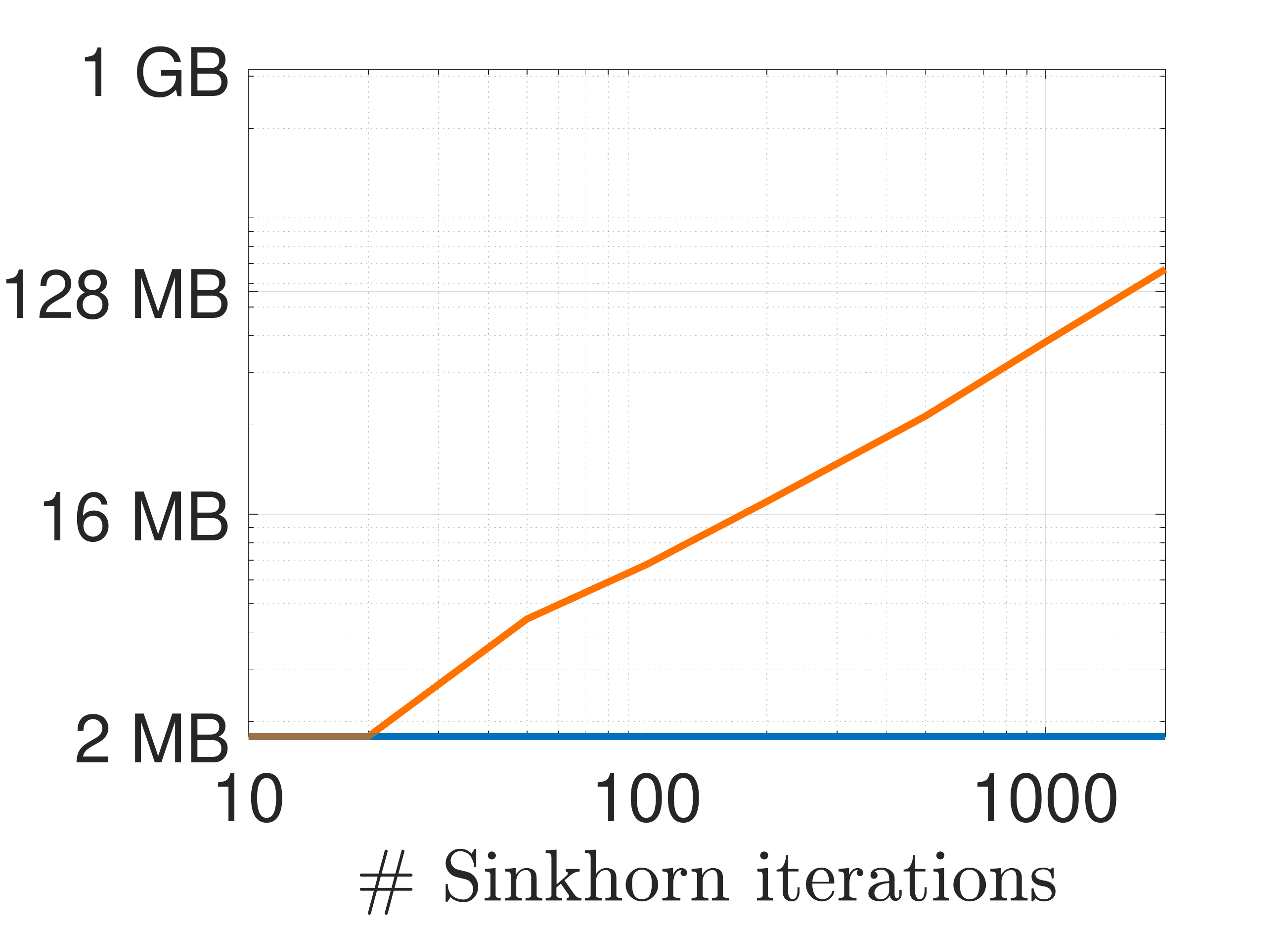}
\includegraphics[width=0.31\textwidth]{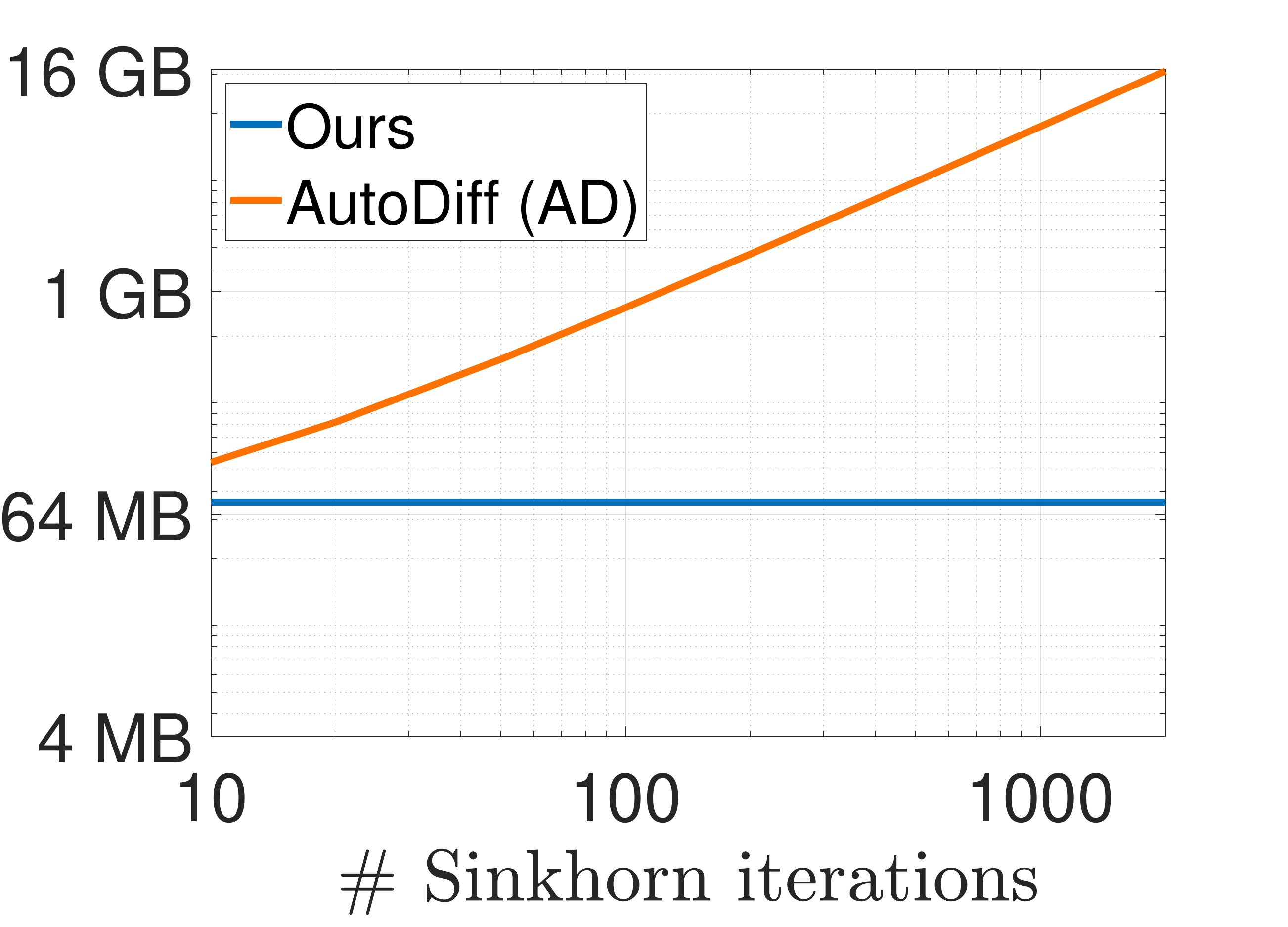}
\end{center}
\vspace{-3pt}
\caption{\textbf{Computational complexity.} We compare the runtime per iteration (top row) and GPU memory requirements (bottom row) of our approach (blue) and automatic differentiation (orange). We consider a broad range of settings with quadratic cost matrices of size $m=n\in\{10,100,1000\}$ and $\tau\in [10,2000]$ Sinkhorn iterations. For the runtime, we show both the total time (solid lines) and the time of only the backward pass (dashed lines). Both ours and AD were implemented in the PyTorch \cite{paszke2019pytorch} framework, where memory is allocated in discrete units, which leads to a large overlap for the minimum allocation size of 2MB (bottom row, left plot).
}
\label{fig:computational_complexity}
\end{figure*}

\begin{figure*}
\vspace{0.05\textwidth}
\hspace{0.08\textwidth}
\begin{overpic}
[width=0.14\textwidth]{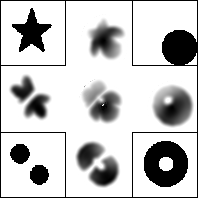}
\put(-65,45){Ours}
\put(28,110){$\tau=10$}
\end{overpic}
\begin{overpic}
[width=0.14\textwidth]{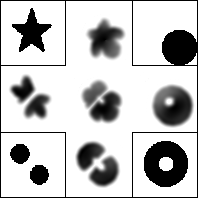}
\put(28,110){$\tau=20$}
\end{overpic}
\begin{overpic}
[width=0.14\textwidth]{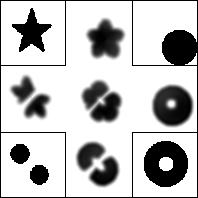}
\put(28,110){$\tau=50$}
\end{overpic}
\begin{overpic}
[width=0.14\textwidth]{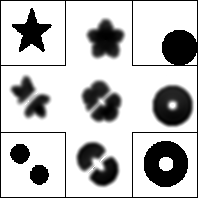}
\put(23,110){$\tau=100$}
\end{overpic}
\begin{overpic}
[width=0.14\textwidth]{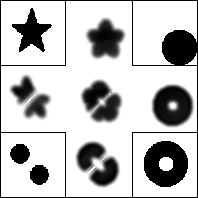}
\put(23,110){$\tau=200$}
\put(77,-70){(OOM)}
\end{overpic}
\begin{overpic}
[width=0.14\textwidth]{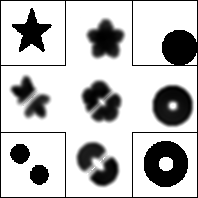}
\put(23,110){$\tau=500$}
\end{overpic}
\\[-0.01\textwidth]

\hspace{0.08\textwidth}
\begin{overpic}
[width=0.14\textwidth]{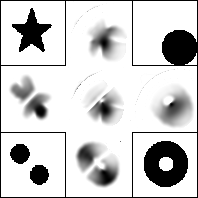}
\put(-65,45){AD}
\end{overpic}
\includegraphics[width=0.14\textwidth]{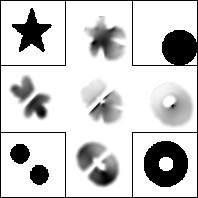}
\includegraphics[width=0.14\textwidth]{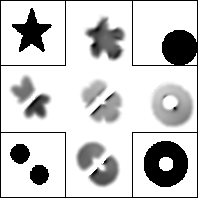}
\includegraphics[width=0.14\textwidth]{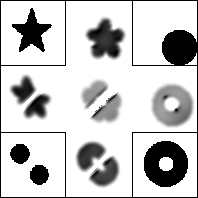}
\caption{\textbf{Wasserstein barycenter.} A comparison between our method (top row) and AD (bottom row) on the application of image barycenter computation. In each cell, we show 5 centroids of 4 input images (corners) with bilinear interpolation weights. The predictions based on the proposed implicit gradients are more stable (providing more crisp interpolations), even for very few Sinkhorn iterations $\tau$. Moreover, AD is out of memory for $\tau\geq 200$. Here, the input images have a resolution of $n=64^2$ and we set $\lambda=0.002$. 
}
\label{fig:img_barycenter}
\end{figure*}

\begin{figure*}
\begin{center}
\scalebox{0.95}{
\hspace{0.09\textwidth}
\begin{overpic}
[width=0.9\textwidth]{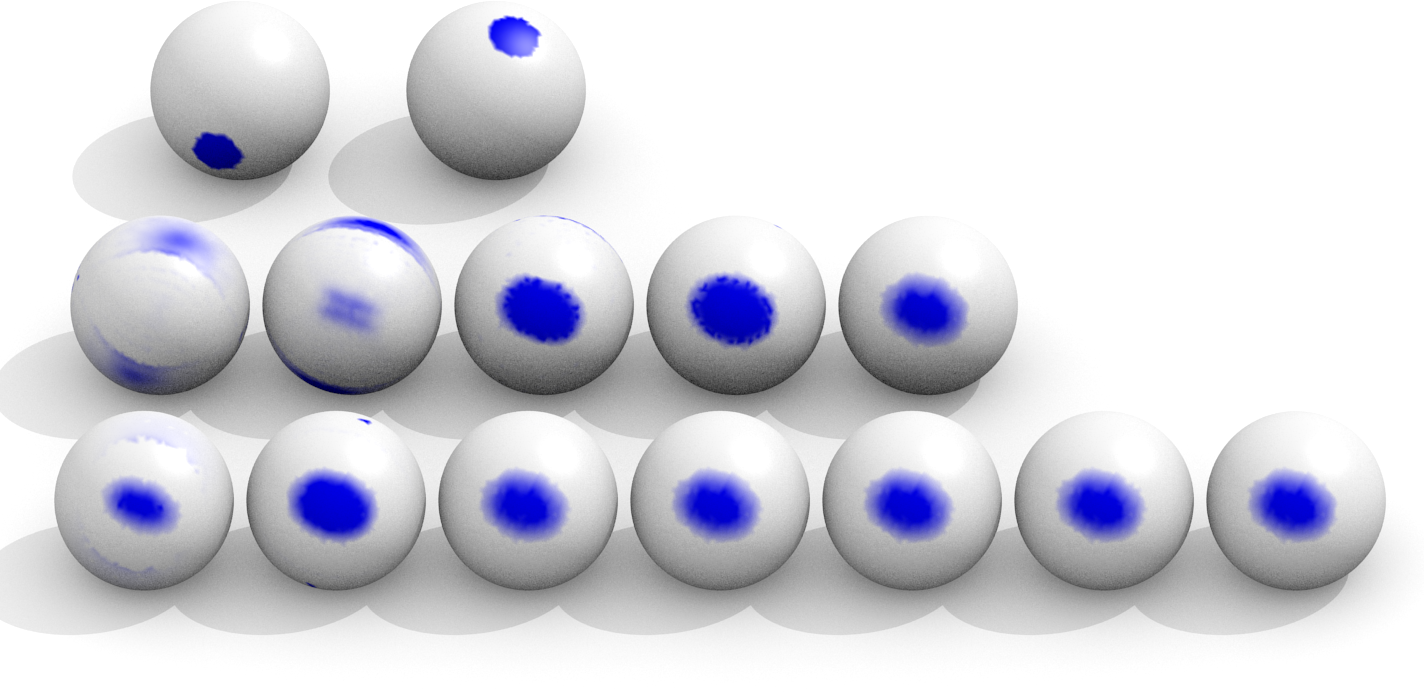}
\put(-8,42){Input pair}
\put(-8,27){AD}
\put(-8,13){Ours}
\put(7,1){$\tau=5$}
\put(22.5,1){$10$}
\put(36.5,1){$20$}
\put(50,1){$50$}
\put(62.5,1){$100$}
\put(76,1){$200$}
\put(86.5,1){$\tau=1000$}
\put(80,26){(OOM)}
\end{overpic}}
\end{center}
\vspace{-10pt}
\caption{\textbf{Manifold barycenter.} We compute barycenters of two circular input distributions on the surface of a sphere (first row). Specifically, we compare the results of minimizing \cref{eq:barycenter} with AD (second row) and implicit gradients (third row). The sphere is discretized as a triangular mesh with $5000$ vertices. On this resolution, AD is out of memory for $\tau\geq 200$ Sinkhorn iterations whereas ours is still feasible for $\tau=1000$. The obtained interpolations produce the slightly elongated shape of an ellipse since the surface of the sphere has a constant positive Gaussian curvature. 
}
\label{fig:manifold_bary}
\end{figure*}

\begin{figure*}
\begin{center}
\includegraphics[width=0.31\textwidth]{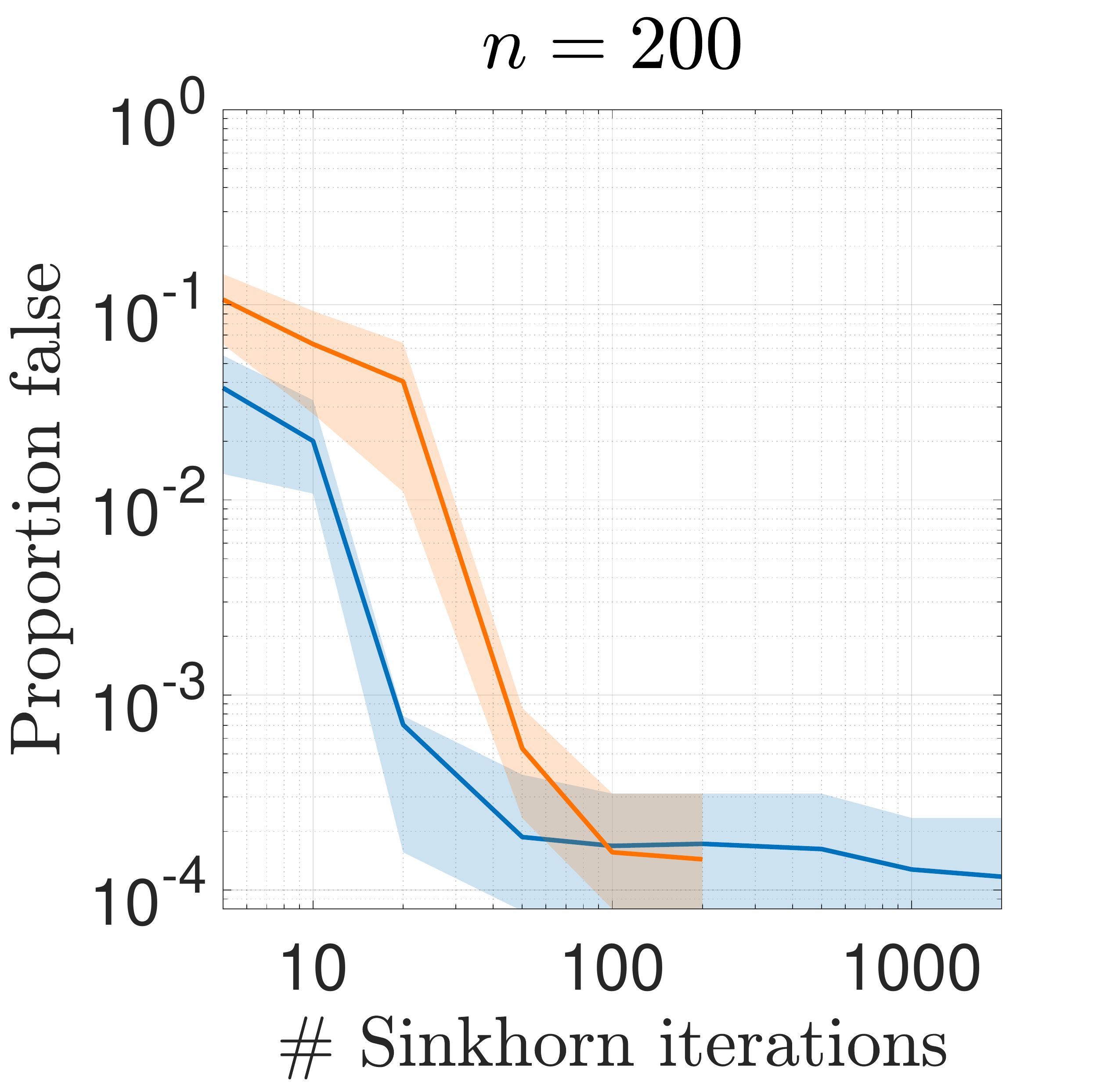}
\includegraphics[width=0.31\textwidth]{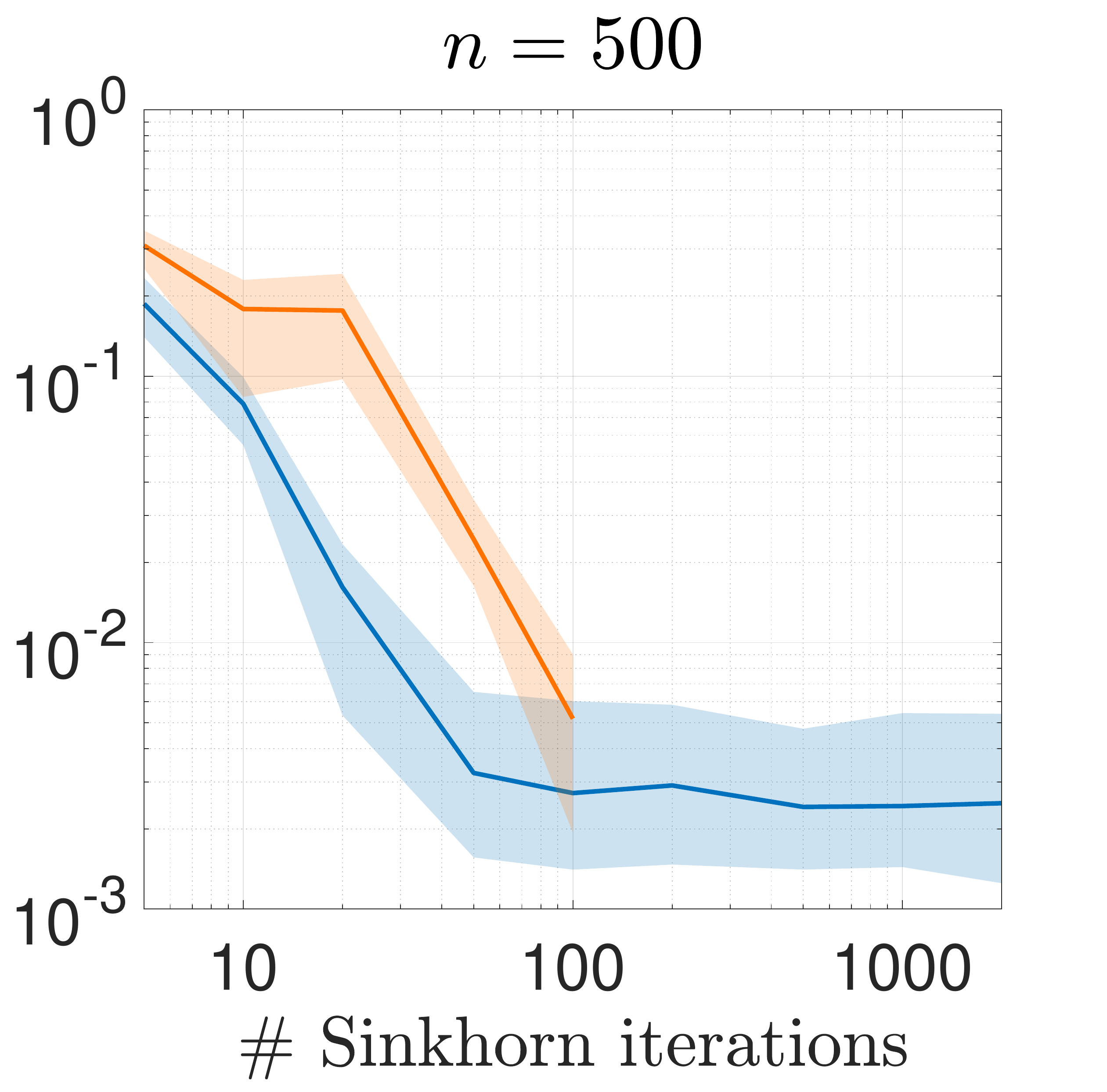}
\begin{overpic}
[width=0.31\textwidth]{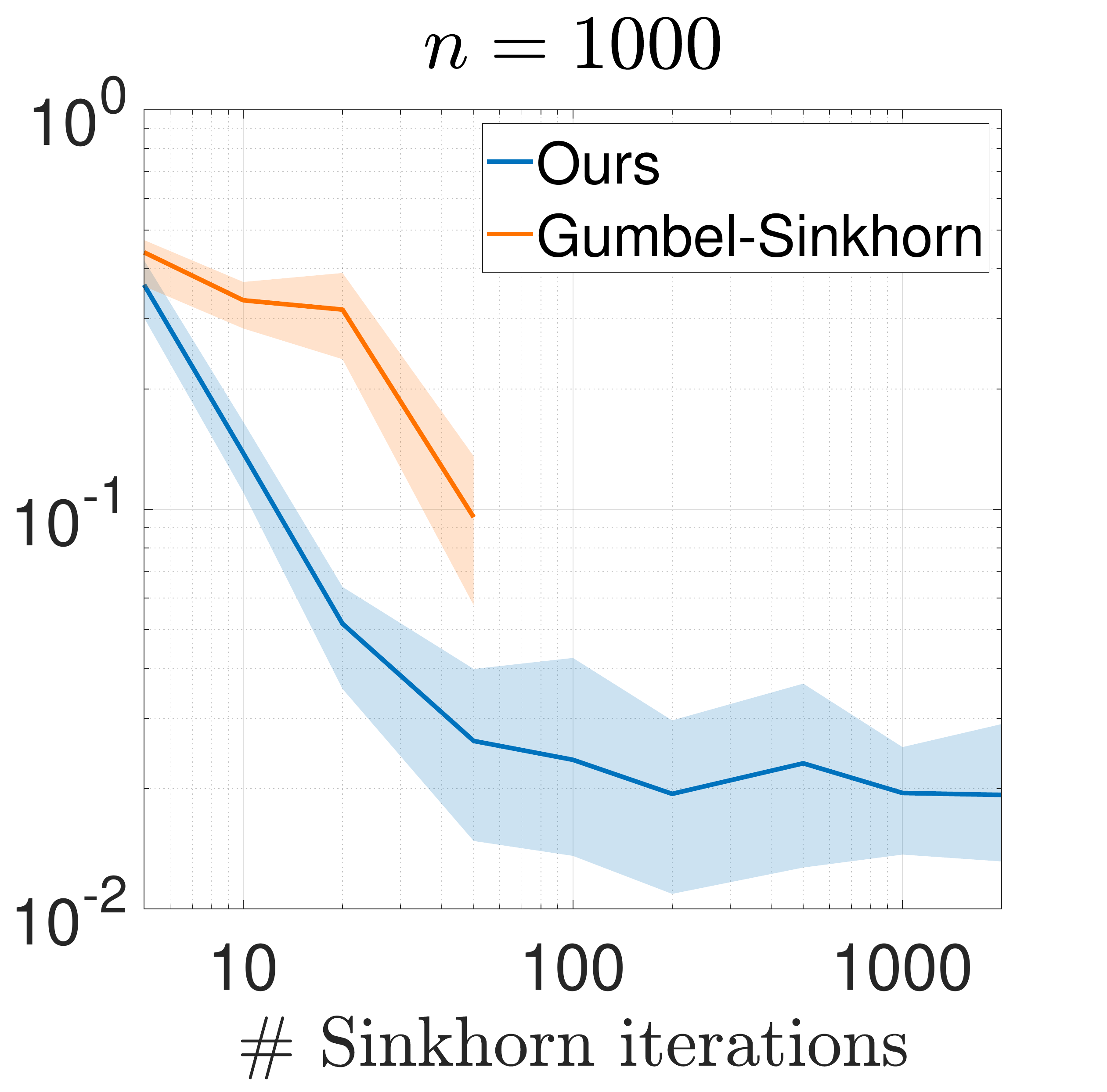}
\put(45,48){\fontsize{7}{8}\selectfont (OOM)}
\end{overpic}
\end{center}
\caption{\textbf{Number sorting.} We show that we can improve the Gumbel-Sinkhorn method \cite{mena2018learning} directly with \cref{alg:backward}. Specifically, we consider the task of permutation learning to sort random number sequences of length $n\in\{200,500,1000\}$, see \cite[Sec~ 5.1]{mena2018learning} for more details. We replace AD in the GS network with implicit differentiation (blue curves) and compare the obtained results to the vanilla GS architecture (orange curves). Our approach yields more accurate permutations while using much less computational resources -- GS is out of memory for $\tau>200,100,50$ forward iterations, respectively. For all settings, we show the mean proportion of correct test set predictions (solid lines), as well as the $10$ and $90$ percentiles (filled areas). The curves are to some degree noisy, since individual results depend on a finite
number of (random) test samples. Also, notice that the log-scale
of the y-axis exaggerates small fluctuations for
$\tau\geq 100$.
}
\label{fig:number_sorting}
\end{figure*}

\begin{table*}
\centering
\resizebox{0.8\linewidth}{!}{
\begin{tabular}{ll|l|lll|lll}
                            &      & \multirow{2}{*}{clean data} & \multicolumn{3}{l|}{\qquad \qquad partial} & \multicolumn{3}{l}{\qquad \qquad \qquad noisy}       \\
 &      &                             & 90\%      & 80\%      & 70\%      & $\sigma=0.001$   & $\sigma=0.01$     & $\sigma=0.1$      \\ \hline
\multirow{2}{*}{Rot. MAE ($\downarrow$)}   & RPM  & \pz0.0299 & 41.1427 & 47.1848 & 52.5945 & 18.5886 & 28.1436 & 43.1884 \\
                            & Ours & \pz0.1371                    & \pz4.4955 & 11.0519 & 20.9274 &\pz 1.0238  & \pz1.2548  & \pz2.2272  \\ \hline
\multirow{2}{*}{Trans. MAE ($\downarrow$)} & RPM  & \pz0.0002                   & \pz0.1743  & \pz0.2126  & \pz0.2490  & \pz0.0848  & \pz0.1187  &\pz 0.1770  \\
                            & Ours & \pz0.0015                   & \pz0.0484  & \pz0.0995  & \pz0.1578  & \pz0.0096  & \pz0.0113  & \pz0.0171  \\ \hline
\multirow{2}{*}{Chamf. dist. ($\downarrow$)}  & RPM  & \pz0.0005                    & \pz4.3413  & \pz4.6829  & \pz4.9581  & \pz2.2077  & \pz3.0492  & \pz4.6935  \\
                            & Ours & \pz0.0054                    & \pz0.5498  & \pz1.4291  & \pz2.2080  & \pz0.0783  & \pz0.1237  & \pz0.4562  \\ \hline
\end{tabular}
}
\caption{\textbf{Point cloud registration.} We compare the quantitative performance of RPM-Net \cite{yew2020rpm} and implicit differentiation on ModelNet40 \cite{wu20153d}. The two architectures are identical except for the altered Sinkhorn module. For all results, we follow the training protocol described in \cite[Sec.~6]{yew2020rpm}. 
Moreover, we assess the ability of the obtained networks to generalize to partial and noisy inputs at test time. For the former, we follow \cite[Sec.~6.6]{yew2020rpm} and remove up to $70\%$ of the input point clouds from a random half-space. For the noisy test set, we add Gaussian white noise $\mathcal{N}(0,\sigma)$ with different variances $\sigma\in\{0.001,0.01,0.1\}$. For all settings, we report the rotation and translation errors, as well as the Chamfer distance to the reference surface. The latter is scaled by a factor of $1e2$ for readability. 
}
\label{table:rpm}
\end{table*}

\begin{figure*}
\begin{center}
\vspace{10pt}
\begin{overpic}
[width=\textwidth]{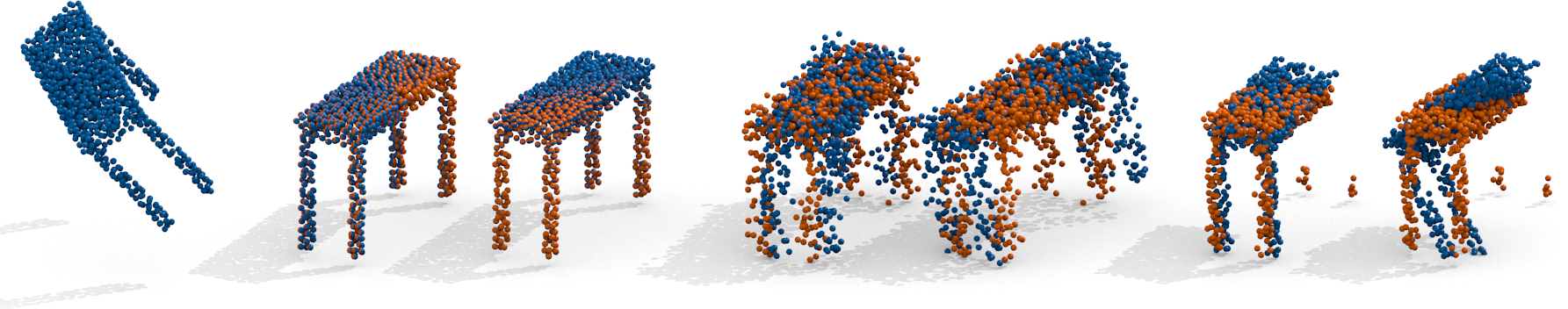}
\put(4,21){Input}
\put(28,21){Clean}
\put(53,21){Noisy $\sigma=0.1$}
\put(82,21){Partial $70\%$}
\put(22,0){\small Ours}
\put(32,0){\small RPM-Net}
\put(51,0){\small Ours}
\put(62,0){\small RPM-Net}
\put(80,0){\small Ours}
\put(90,0){\small RPM-Net}
\end{overpic}
\\[2pt]

\begin{overpic}
[width=\textwidth]{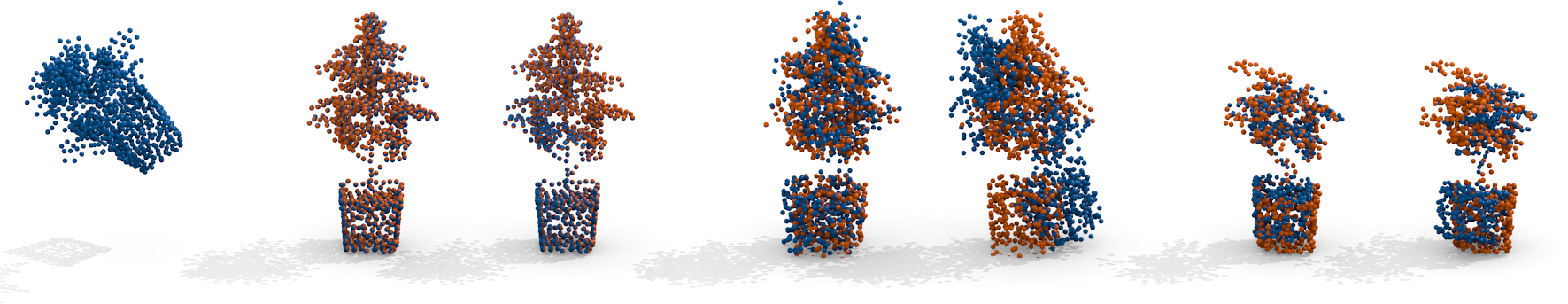}
\put(22,0){\small Ours}
\put(32,0){\small RPM-Net}
\put(51,0){\small Ours}
\put(62,0){\small RPM-Net}
\put(80,0){\small Ours}
\put(90,0){\small RPM-Net}
\end{overpic}
\end{center}
\vspace{-5pt}
\caption{\textbf{Point cloud registration.} 
Qualitative comparisons of RPM-Net \cite{yew2020rpm} and the improved version based on implicit differentiation. In each row, we show a different test pair with the input pose $X$ (1st column, blue), as well as the overlap of the reference pose $Y$ (orange) and the predicted pose (blue) for the clean, noisy, and partial settings. Both approaches work well for the clean data, but ours generalizes more robustly to noisy and partial pairs.
}
\label{fig:rpm_qual_1}
\end{figure*}

\section{Experiments}\label{sec:experiments}

In \cref{subsec:computationalcomplexity}, we empirically compare the computation cost of \cref{alg:backward} to automatic differentiation (AD).
In \cref{subsec:barycenter} and \cref{subsec:permutation}, we show results on two common classes of applications where we want to learn the marginals $\va$ and the cost matrix $\mC$ respectively.
We assume a fixed GPU memory (VRAM) budget of 24GB -- any setting that exceeds this limit is deemed out of memory (OOM). 

\subsection{Computation cost}\label{subsec:computationalcomplexity}

We empirically compare the computation cost of our algorithm with the standard automatic differentiation approach, see \cref{fig:computational_complexity}. All results were computed on a single NVIDIA Quadro RTX 8000 graphics card. In practice, the computation cost of both approaches primarily depends on the parameters $m,n,\tau$. It is for the most part indifferent to other hyperparameters and the actual values of $\mC,\va,\vb$. We therefore use random (log normal distributed) cost matrices $\ln{\mC_{i,j}}\sim\mathcal{N}(0,1)$ and uniform marginals $\va=\vb=\frac{1}{n}\bbone_n$ with $m=n\in\{10,100,1000\}$. For each setting, we report the cost of the forward and backward pass averaged over 1k iterations.
Depending on $m,n$, our approach is faster for $\tau\gtrsim  40,50,90$ iterations. Note that our backward pass is independent of the number of forward iterations $\tau$. Finally, the memory requirements are dramatically higher for AD, since it needs to maintain the computation graph of all $\tau$ forward iterations. In practice, this often limits the admissible batch size or input resolution, see \cref{subsec:barycenter} and \cref{subsec:permutation}.

\subsection{Wasserstein barycenters}\label{subsec:barycenter}

The main idea of Barycenter computation is to interpolate between a collection of objects $\{\vb_1,\dots,\vb_k\}\subset\mathbb{R}^n$ as a convex combination with weights that lie on the probability simplex $\vw\in\Delta_k$, see \cref{eq:probsimplex}. Specifically, we optimize
\begin{align}\label{eq:barycenter}
    \va^*:=&\underset{\va\in\Delta_n}{\arg\min}\sum_{i=1}^kw_id(\va,\vb_i)\qquad\text{with}\\\label{eq:barycenterb}
    d(\va,\vb):=&\min_{\mP\in\Pi(\va,\vb)}\langle\mP,\mD\rangle_F-\lambda h(\mP),
\end{align}
where $\mD\in\mathbb{R}^{n\times n}$ denotes the squared pairwise distance matrix between the domains of $\va$ and $\vb$.
We use the Adam optimizer \cite{kingma2014adam} for the outer optimization in \cref{eq:barycenter}. The inner optimization \cref{eq:barycenterb} is a special case of \cref{eq:sinkhornoperator}. 
Overall, \cref{eq:barycenter} allows us to compute geometrically meaningful interpolations in arbitrary metric spaces. We consider the explicit tasks of interpolating between images in \cref{fig:img_barycenter} and functions on manifolds in \cref{fig:manifold_bary}. 
Note that there are a number of specialized algorithms that minimize \cref{eq:barycenter} in a highly efficient manner \cite{cuturi2014fast,solomon2015convolutional,luise2019sinkhorn}. 
In \cref{subsec:mnistimageclustering}, we further show how to apply the barycenter technique to image clustering on the MNIST dataset. 

\subsection{Permutation learning and matching}\label{subsec:permutation}
\paragraph{Number sorting.} The Sinkhorn operator is nowadays a standard tool to parameterize approximate permutations within a neural network. 
One work that clearly demonstrates the effectiveness of this approach is the Gumbel-Sinkhorn (GS) method \cite{mena2018learning}. 
The main idea is to learn the natural ordering of sets of input elements $\{x_1,\dots,x_n\}$, see \cref{subsec:additionalpermutation} for more details. Here, we consider the concrete example of learning to sort real numbers from the unit interval $x_i\in[0,1]$ for $n\in\{200,500,1000\}$ numbers. We compare the implicit Sinkhorn module to the vanilla GS method in \cref{fig:number_sorting}. Without further modifications, our method significantly decreases the error at test time, defined as the proportion of incorrectly sorted elements.

\paragraph{Point cloud registration.} Several recent methods use the Sinkhorn operator as a differentiable, bijective matching layer for deep learning \cite{sarlin2020superglue,yew2020rpm,yang2020mapping,liu2020learning,eisenberger2020deep}. 
Here, we consider the concrete application of rigid point cloud registration \cite{yew2020rpm} and show that we can improve the performance with implicit differentiation, see \cref{table:rpm}. While our results on the clean test data are comparable but slightly worse than the vanilla RPM-Net \cite{yew2020rpm}, our module generalizes more robustly to partial and noisy observations. This indicates that, since computing gradients with our method is less noisy than AD, it helps to learn a robust matching policy that is overall more consistent, see \cref{fig:rpm_qual_1} for qualitative comparisons. We provide further details on the RPM-Net baseline and more qualitative results in \cref{subsec:additionalpermutation}.

\section{Conclusion}
We presented a unifying framework that provides analytical gradients of the Sinkhorn operator in its most general form.
In contrast to more specialized approaches~\cite{luise2018differential,flamary2018wasserstein,campbell2020solving,cuturi2020supervised,klatt2020empirical}, our algorithm can be deployed in a broad range of applications in a straightforward manner.
Choosing the number of Sinkhorn iterations $\tau\in\mathbb{N}$ is generally subject to a trade-off between the computation cost and accuracy. The main advantage of implicit differentiation is that it proves to be much more scalable than AD, since the backward pass is independent of $\tau$. 
Our experiments demonstrate that combining the implicit Sinkhorn module with existing approaches often improves the performance. 
We further provide theoretical insights and error bounds that quantify the accuracy of \cref{alg:backward} for noisy inputs.

\paragraph{Limitations \& societal impact}
In our view, one of the main limitations of \cref{alg:backward} is that AD results in a faster training time for very few iterations $\tau\approx 10$. Whether this is offset by the empirically more stable training (see \cref{subsec:barycenter} and \cref{subsec:permutation}) has to be judged on a case-by-case basis. In terms of the societal impact, one of the major advantages of our method is that it reduces computation time and GPU memory demand of Sinkhorn layers within neural networks. It thereby has the potential to make such techniques more accessible to individuals and organizations with limited access to computational resources. 

\section*{Acknowledgements}
This work was supported by the ERC Advanced Grant SIMULACRON and the Munich School for Data Science.

{\small
\bibliographystyle{ieee_fullname}
\bibliography{refs}
}

\clearpage
\appendix
\definecolor{dkgreen}{rgb}{0,0.6,0}
\definecolor{gray}{rgb}{0.5,0.5,0.5}
\definecolor{mauve}{rgb}{0.58,0,0.82}

\lstset{frame=tb,
  language=Python,
  aboveskip=3mm,
  belowskip=3mm,
  showstringspaces=false,
  columns=flexible,
  basicstyle={\small\ttfamily},
  numbers=none,
  numberstyle=\tiny\color{gray},
  keywordstyle=\color{blue},
  commentstyle=\color{dkgreen},
  stringstyle=\color{mauve},
  breaklines=true,
  breakatwhitespace=true,
  tabsize=3
}

\begin{figure*}[h]
\begin{lstlisting}

import torch

class Sinkhorn(torch.autograd.Function):
    @staticmethod
    def forward(ctx, c, a, b, num_sink, lambd_sink):
        log_p = -c / lambd_sink
        log_a = torch.log(a).unsqueeze(dim=1)
        log_b = torch.log(b).unsqueeze(dim=0)
        for _ in range(num_sink):
            log_p -= (torch.logsumexp(log_p, dim=0, keepdim=True) - log_b)
            log_p -= (torch.logsumexp(log_p, dim=1, keepdim=True) - log_a)
        p = torch.exp(log_p)

        ctx.save_for_backward(p, torch.sum(p, dim=1), torch.sum(p, dim=0))
        ctx.lambd_sink = lambd_sink
        return p

    @staticmethod
    def backward(ctx, grad_p):
        p, a, b = ctx.saved_tensors
        m, n = p.shape

        grad_p *= -1 / ctx.lambd_sink * p
        K = torch.cat((
            torch.cat((torch.diag(a), p), dim=1), 
            torch.cat((p.T, torch.diag(b)), dim=1)), 
            dim=0)[:-1, :-1]
        t = torch.cat((
            grad_p.sum(dim=1), 
            grad_p[:, :-1].sum(dim=0)), 
            dim=0).unsqueeze(1)
        grad_ab, _ = torch.solve(t, K)
        grad_a = grad_ab[:m, :]
        grad_b = torch.cat((grad_ab[m:, :], torch.zeros([1, 1], device=device, dtype=torch.float32)), dim=0)
        U = grad_a + grad_b.T
        grad_p -= p * U
        grad_a = -ctx.lambd_sink * grad_a.squeeze(dim=1)
        grad_b = -ctx.lambd_sink * grad_b.squeeze(dim=1)

        return grad_p, grad_a, grad_b, None, None
        
\end{lstlisting}
\caption{A PyTorch implementation of \cref{alg:backward}.
}
\label{alg:pytorch}
\end{figure*}
\section{PyTorch implementation}\label{sec:pytorchimplementation}
Our proposed algorithm can be easily integrated into existing deep learning architectures. Here, we provide our PyTorch \cite{paszke2019pytorch} implementation of this module, see \cref{alg:pytorch}. The forward pass is the standard Sinkhorn matrix scaling algorithm \cite{cuturi2013sinkhorn} with a robust log-space implementation. The backward function contains an implementation of our \cref{alg:backward}.

\section{Additional experiments}\label{sec:additionalexperiments}

\subsection{Gradient accuracy}\label{subsec:gradientaccuracy}

\begin{figure*}
\begin{center}
\hspace{0.06\textwidth}
\begin{overpic}
[width=0.23\textwidth]{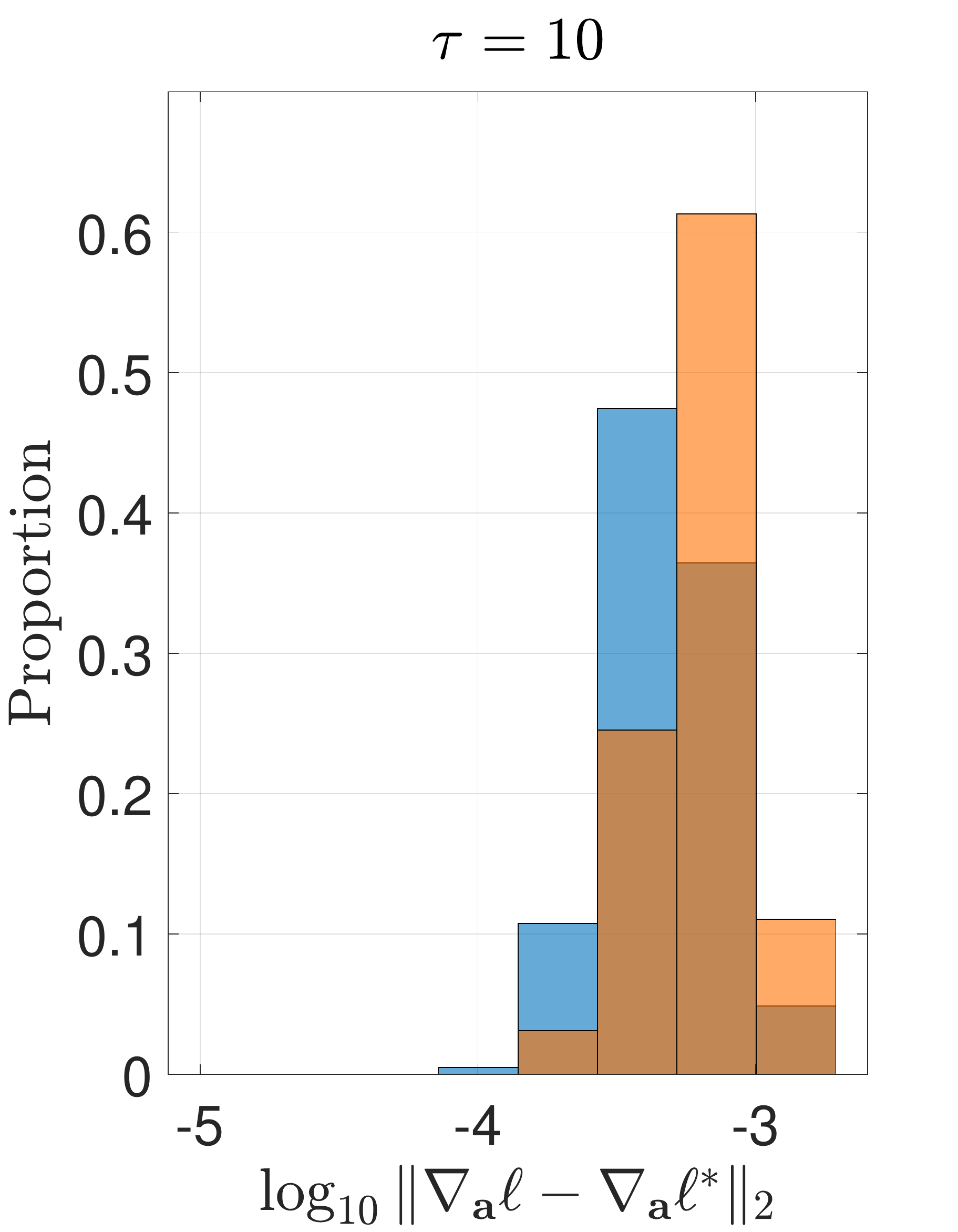}
\put(-10,26){{\rotatebox{90}{Image barycenter}}}
\end{overpic}
\includegraphics[width=0.23\textwidth]{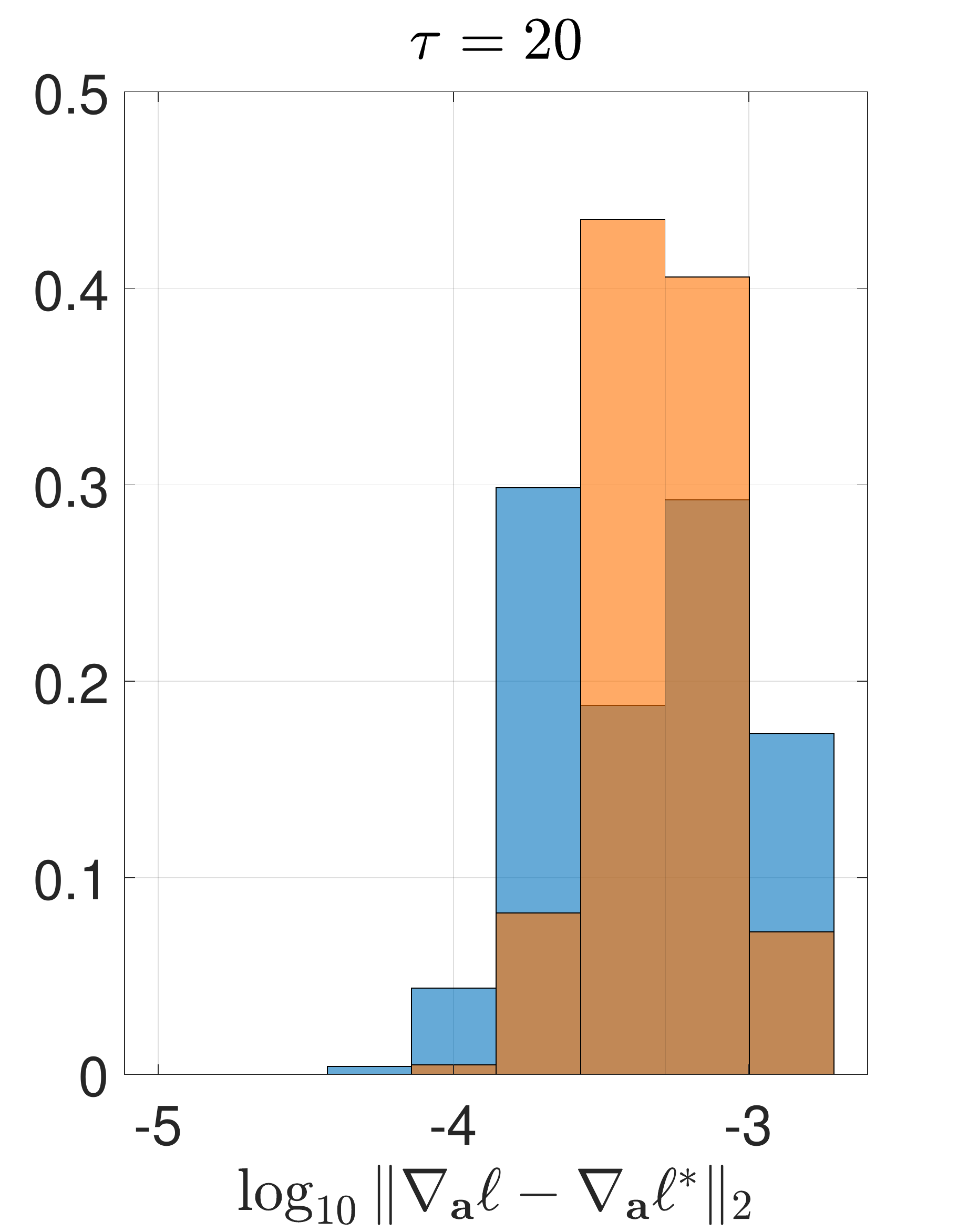}
\includegraphics[width=0.23\textwidth]{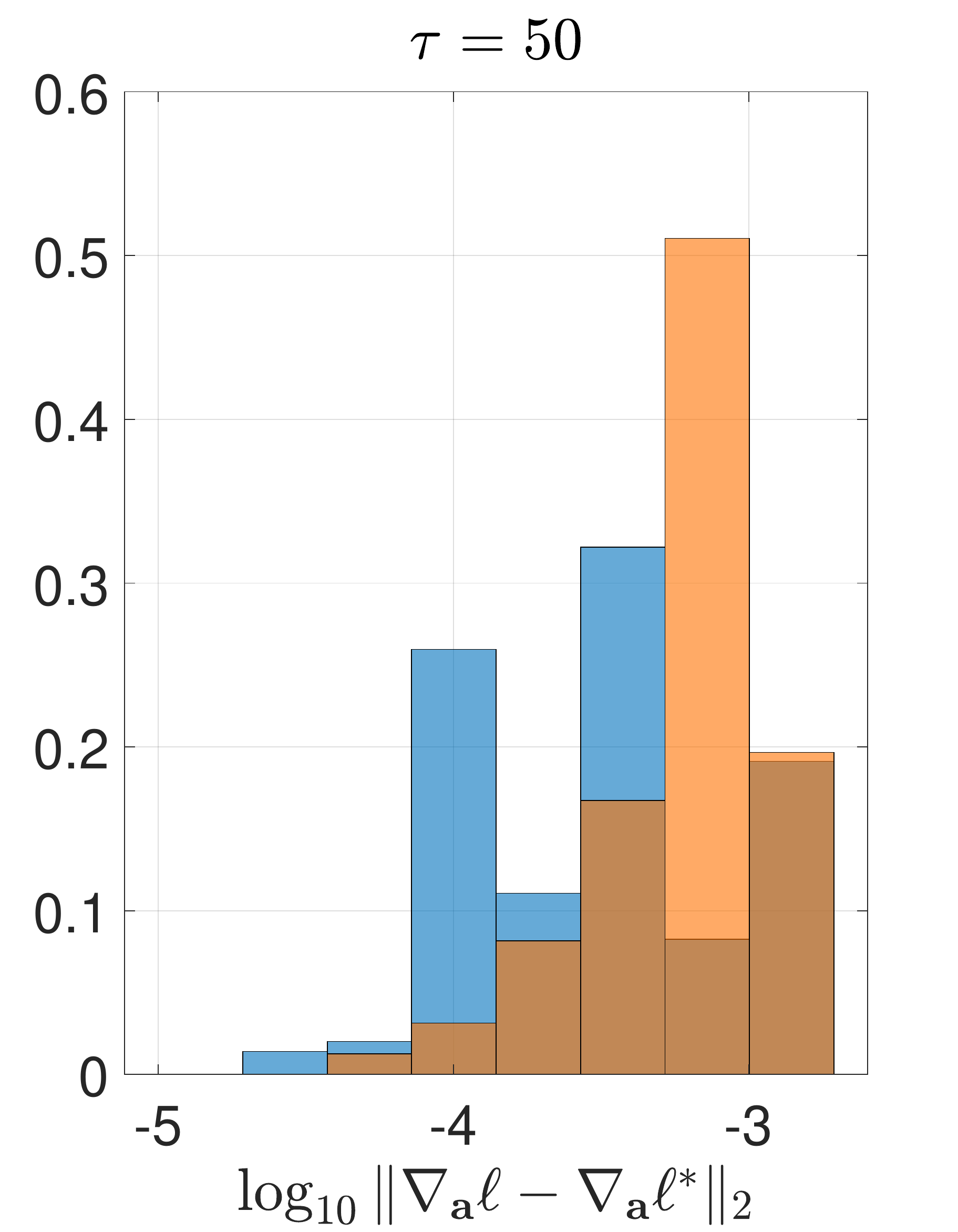}
\includegraphics[width=0.23\textwidth]{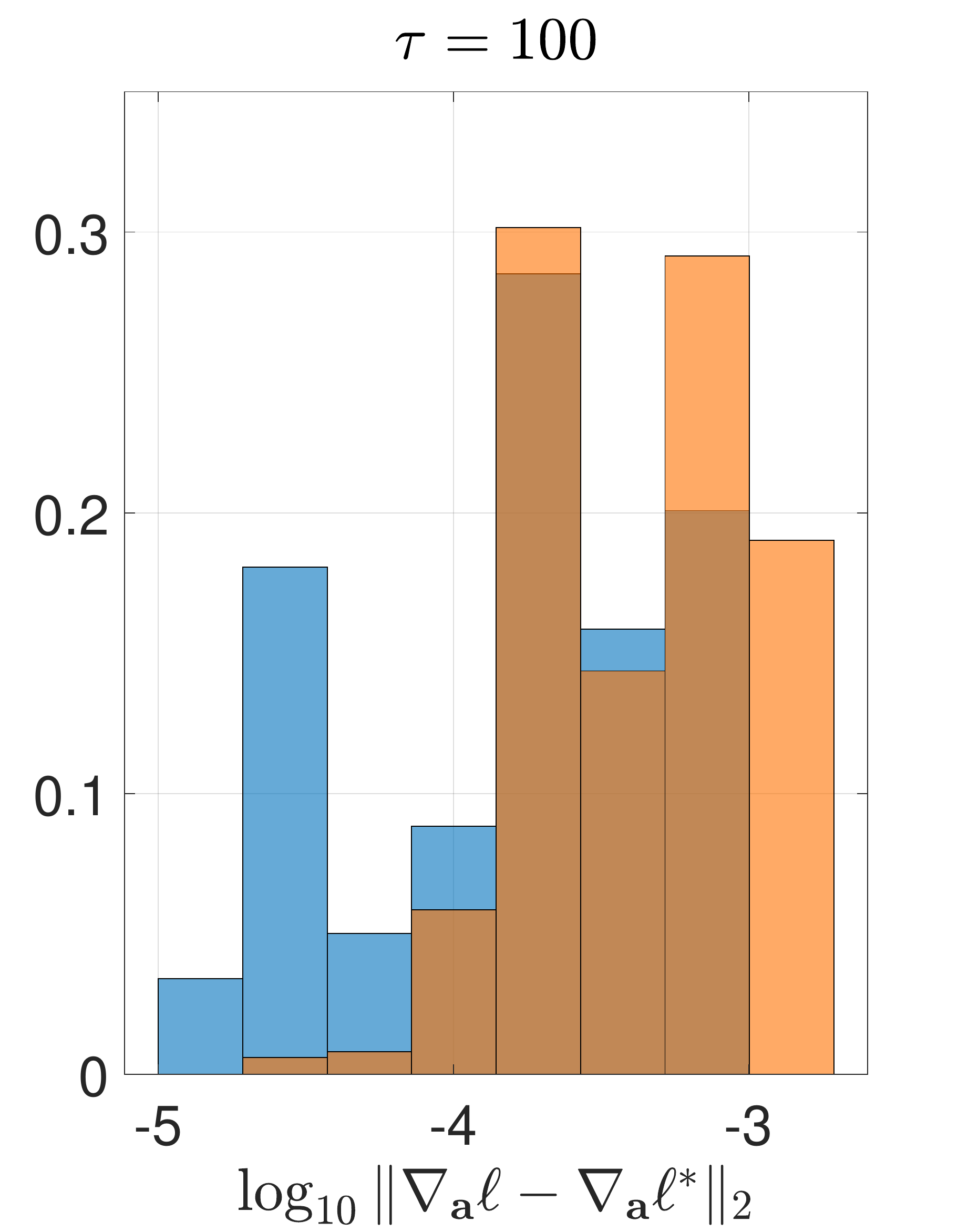}

\hspace{0.06\textwidth}
\begin{overpic}
[width=0.23\textwidth]{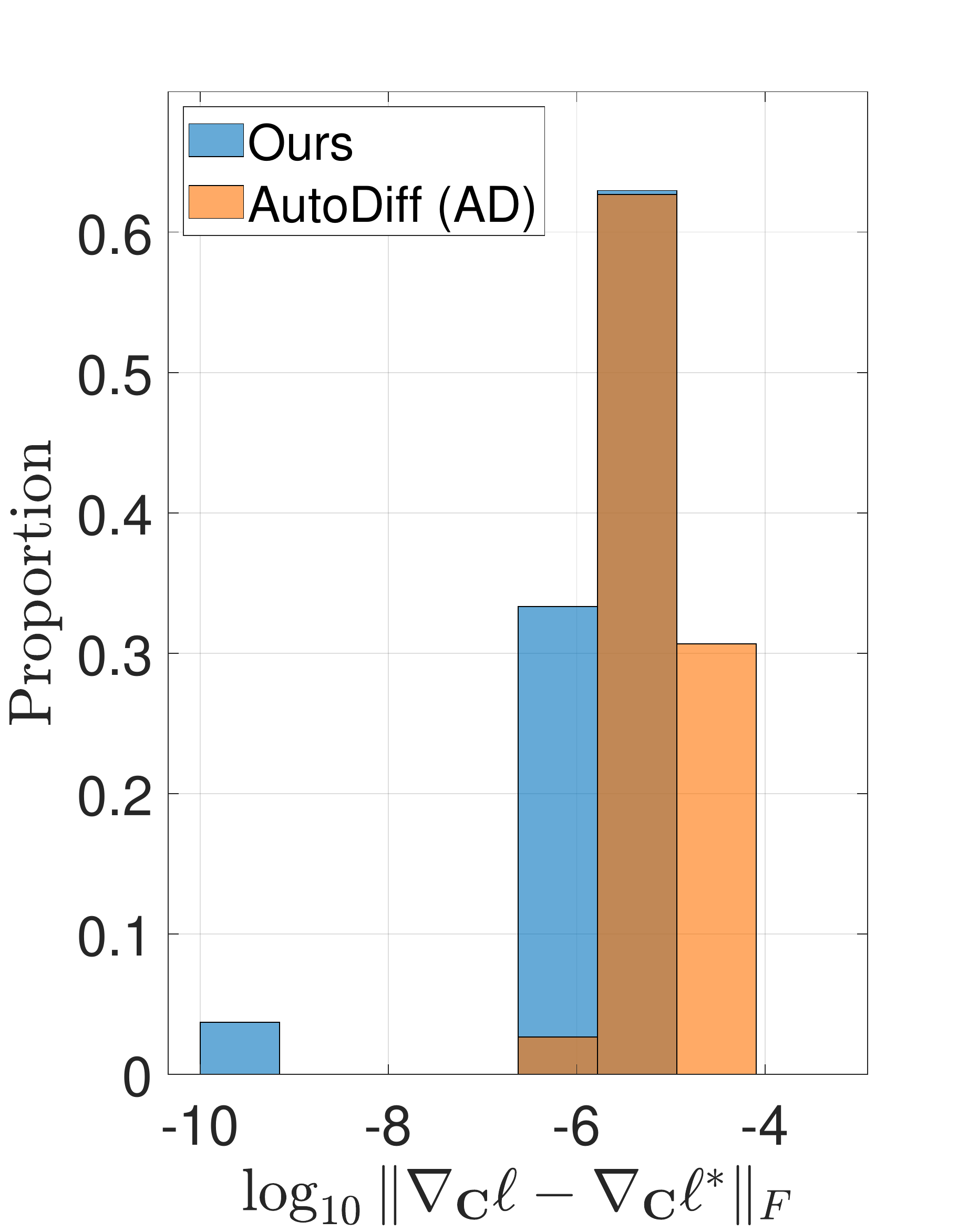}
\put(-10,33){{\rotatebox{90}{Number sorting}}}
\end{overpic}
\includegraphics[width=0.23\textwidth]{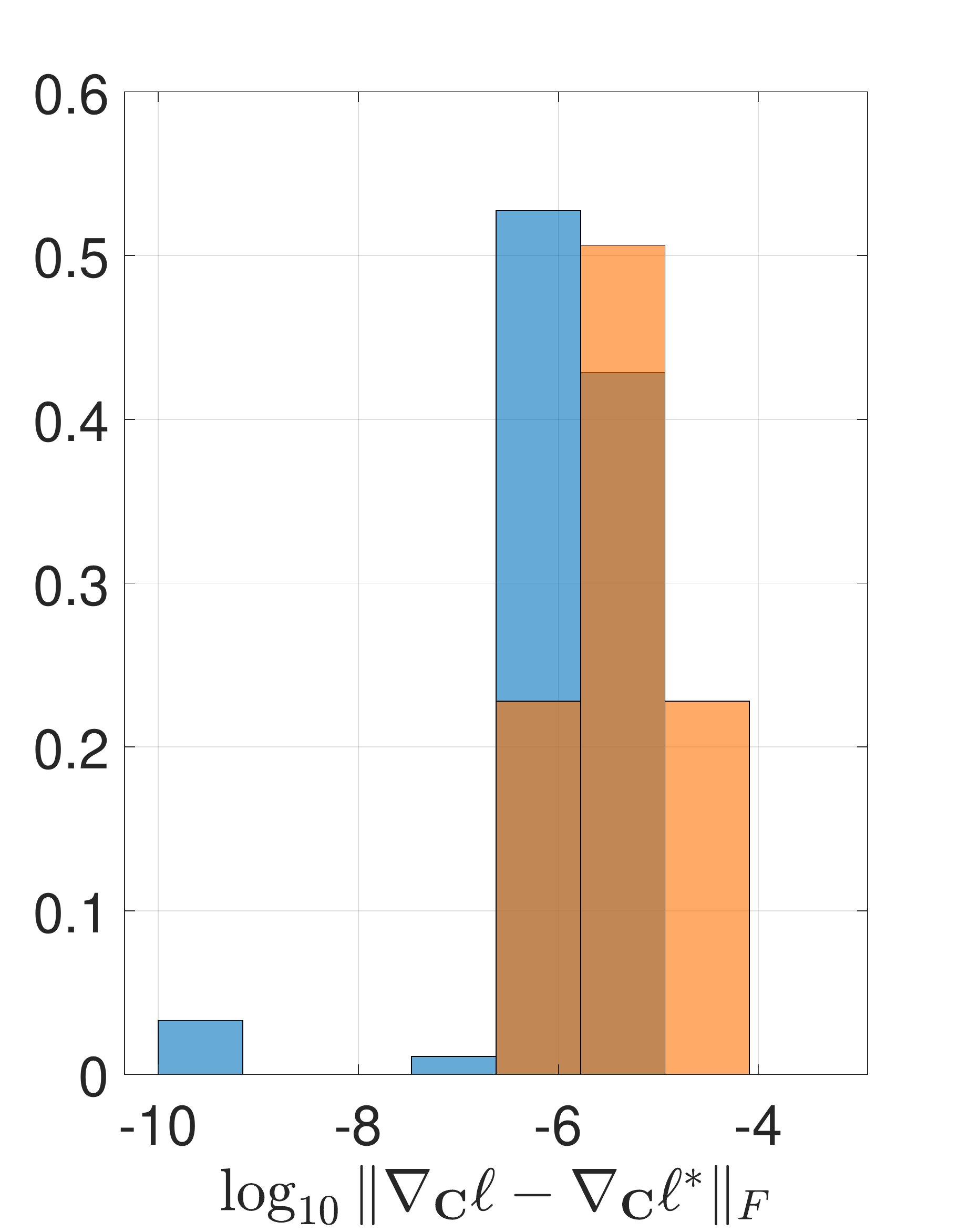}
\includegraphics[width=0.23\textwidth]{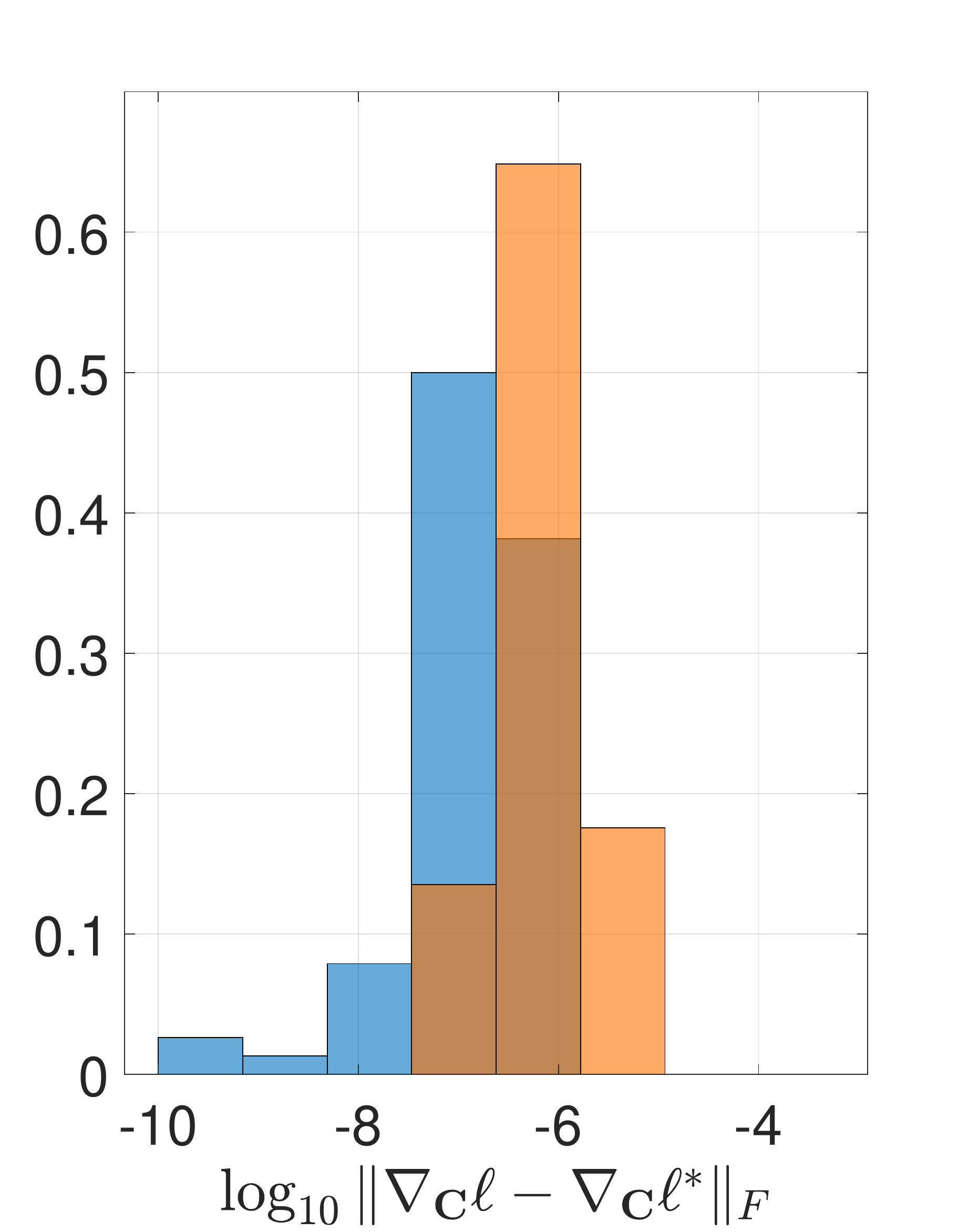}
\includegraphics[width=0.23\textwidth]{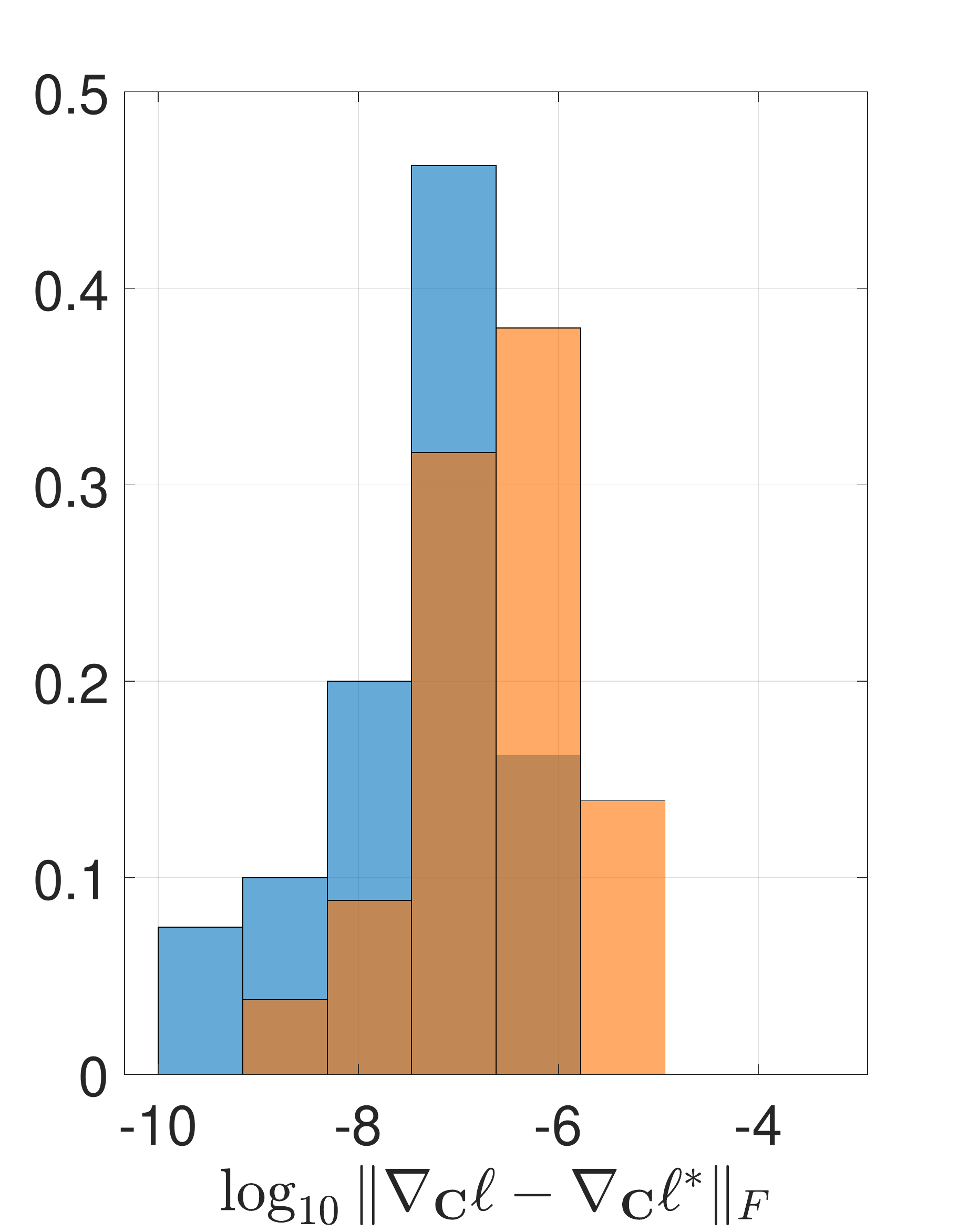}
\end{center}
\caption{\textbf{Gradient accuracy.} We empirically assess the accuracy of the error bound discussed in \cref{thm:errorbounds}. Specifically, we show the accuracy of the gradients $\nabla_{\va}\ell$ for the image barycenter experiment in \cref{subsec:barycenter} (top row) and $\nabla_{\mC}\ell$ for the number sorting experiment in \cref{subsec:permutation} (bottom row). While both distributions have a large overlap, the gradients from our approach (blue) are noticeably more accurate than AD (orange) on average. Note that all comparisons show histograms on a log scale. 
}
\label{fig:gradient_approx}
\end{figure*}

\begin{figure*}
\begin{center}

\hspace{\wbary\textwidth}
\begin{overpic}
[width=\wbary\textwidth]{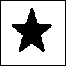}
\put(-70,45){Input}
\end{overpic}
\includegraphics[width=\wbary\textwidth]{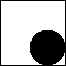}
\hspace{0.63\textwidth}

\vspace{0.02\linewidth}
\hspace{\wbary\textwidth}
\begin{overpic}
[width=\wbary\textwidth]{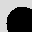}
\put(-123,45){Barycenter}
\end{overpic}
\includegraphics[width=\wbary\textwidth]{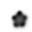}
\includegraphics[width=\wbary\textwidth]{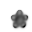}
\includegraphics[width=\wbary\textwidth]{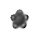}
\includegraphics[width=\wbary\textwidth]{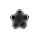}
\includegraphics[width=\wbary\textwidth]{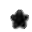}
\includegraphics[width=\wbary\textwidth]{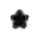}
\includegraphics[width=\wbary\textwidth]{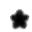}

\hspace{\wbary\textwidth}
\begin{overpic}
[width=\wbary\textwidth]{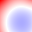}
\put(-128,45){$\nabla_{\va}\ell$ -- Ours}
\end{overpic}
\includegraphics[width=\wbary\textwidth]{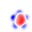}
\includegraphics[width=\wbary\textwidth]{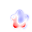}
\includegraphics[width=\wbary\textwidth]{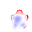}
\includegraphics[width=\wbary\textwidth]{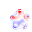}
\includegraphics[width=\wbary\textwidth]{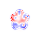}
\includegraphics[width=\wbary\textwidth]{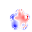}
\includegraphics[width=\wbary\textwidth]{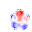}

\hspace{\wbary\textwidth}
\begin{overpic}
[width=\wbary\textwidth]{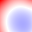}
\put(-128,45){$\nabla_{\va}\ell$ -- AD}
\end{overpic}
\includegraphics[width=\wbary\textwidth]{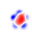}
\includegraphics[width=\wbary\textwidth]{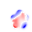}
\includegraphics[width=\wbary\textwidth]{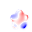}
\includegraphics[width=\wbary\textwidth]{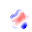}
\includegraphics[width=\wbary\textwidth]{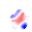}
\includegraphics[width=\wbary\textwidth]{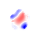}
\includegraphics[width=\wbary\textwidth]{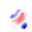}

\hspace{0.093\textwidth}
\begin{overpic}
[width=\wbary\textwidth]{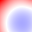}
\put(-128,45){$\nabla_{\va}\ell$ -- g.t.}
\put(27,-28){$t=0$}
\end{overpic}
\begin{overpic}
[width=\wbary\textwidth]{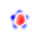}
\put(38,-28){$10$}
\end{overpic}
\begin{overpic}
[width=\wbary\textwidth]{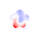}
\put(38,-28){$20$}
\end{overpic}
\begin{overpic}
[width=\wbary\textwidth]{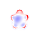}
\put(38,-28){$30$}
\end{overpic}
\begin{overpic}
[width=\wbary\textwidth]{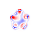}
\put(38,-28){$40$}
\end{overpic}
\begin{overpic}
[width=\wbary\textwidth]{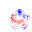}
\put(38,-28){$50$}
\end{overpic}
\begin{overpic}
[width=\wbary\textwidth]{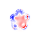}
\put(38,-28){$60$}
\end{overpic}
\begin{overpic}
[width=\wbary\textwidth]{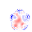}
\put(23,-28){$t=70$}
\end{overpic}
\\[0.03\linewidth]
\end{center}
\caption{\textbf{Image barycenter gradients.} A qualitative comparison of our gradients (3rd row), the AD gradients (4th row), and the ground truth gradients (last row) for the image barycenter experiment from \cref{subsec:barycenter}. Specifically, we consider the task of interpolating between two input images (1st row) with uniform interpolation weights $w_1=w_2=0.5$. We show intermediate snapshots of the obtained barycenter image (2nd row) for different numbers of gradient descent iterations $t\in\{0,\dots,70\}$ that result from minimizing the energy in \Eqref{eq:barycenter}.
}
\label{fig:gradient_approx_qual}
\end{figure*}

\begin{figure*}
\begin{center}
\vspace{0.05\linewidth}
\begin{overpic}
[width=0.17\textwidth]{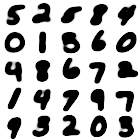}
\put(32,115){$\tau=5$}
\put(-60,45){Ours}
\end{overpic}
\hspace{0.02\linewidth}
\begin{overpic}
[width=0.17\textwidth]{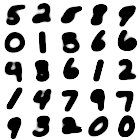}
\put(30,115){$\tau=10$}
\end{overpic}
\hspace{0.02\linewidth}
\begin{overpic}
[width=0.17\textwidth]{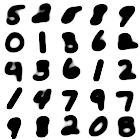}
\put(30,115){$\tau=50$}
\end{overpic}
\hspace{0.02\linewidth}
\begin{overpic}
[width=0.17\textwidth]{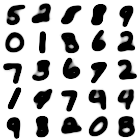}
\put(28,115){$\tau=100$}
\end{overpic}
\\[0.02\linewidth]

\begin{overpic}
[width=0.17\textwidth]{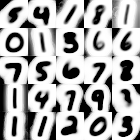}
\put(-60,45){AD}
\end{overpic}
\hspace{0.02\linewidth}
\includegraphics[width=0.17\textwidth]{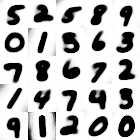}
\hspace{0.02\linewidth}
\includegraphics[width=0.17\textwidth]{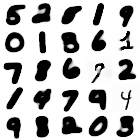}
\hspace{0.02\linewidth}
\includegraphics[width=0.17\textwidth]{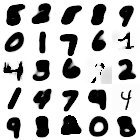}
\end{center}
\caption{\textbf{MNIST k-means clustering}. A comparison of implicit differentiation and AD on the task of k-means image clustering. For both approaches, we show the predicted $k=25$ clusters for $\tau\in\{5,10,50,100\}$ Sinkhorn iterations. We choose more than $10$ clusters to capture several different appearances and styles for each digit. To make individual results more comparable, we use an identical random initialization of the cluster centroids for all settings. For AD, the maximum permissible batch sizes for the 4 settings are $512,256,64,32$, whereas the implicit differentiation algorithm consistently allows for a batch size of $1024$.
}
\label{fig:imageclustering}
\end{figure*}

\begin{figure*}
\begin{center}
\vspace{0.02\linewidth}
\begin{overpic}
[width=\textwidth]{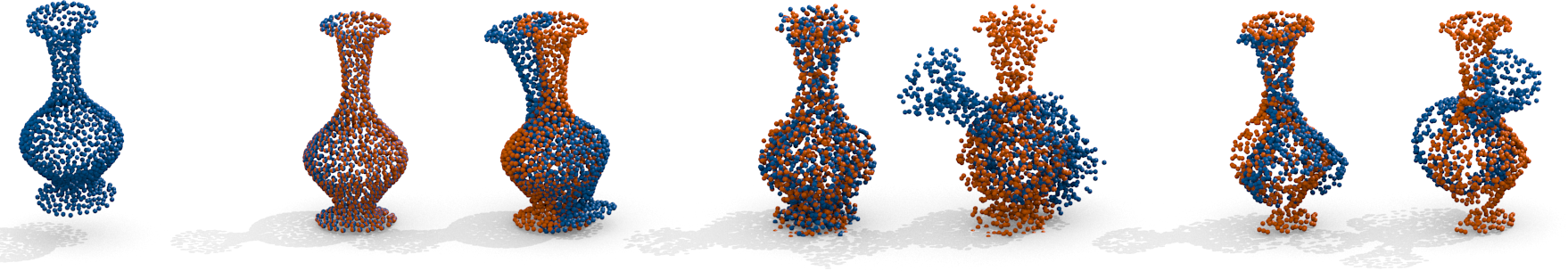}
\put(4,20){Input}
\put(28,20){Clean}
\put(53,20){Noisy $\sigma=0.1$}
\put(82,20){Partial $70\%$}
\put(20,-2){\small Ours}
\put(30,-2){\small RPM-Net}
\put(50,-2){\small Ours}
\put(60,-2){\small RPM-Net}
\put(80,-2){\small Ours}
\put(90,-2){\small RPM-Net}
\end{overpic}
\\[0.05\linewidth]

\begin{overpic}
[width=\textwidth]{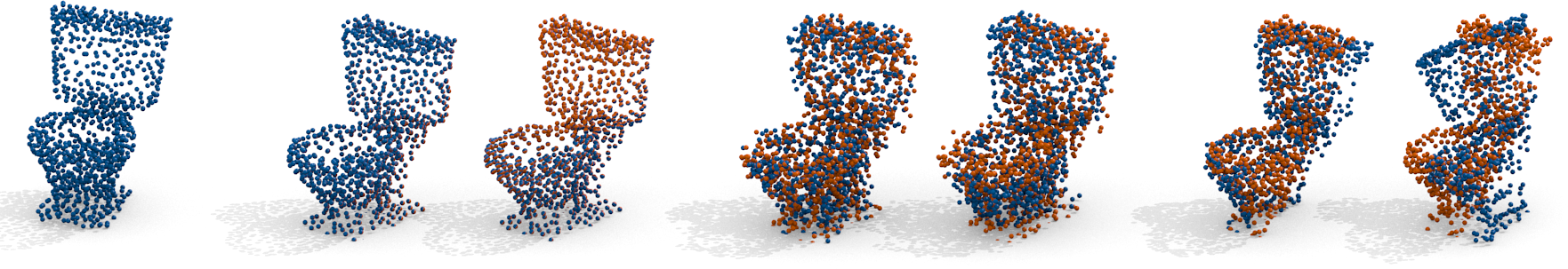}
\put(20,-2){\small Ours}
\put(30,-2){\small RPM-Net}
\put(50,-2){\small Ours}
\put(60,-2){\small RPM-Net}
\put(80,-2){\small Ours}
\put(90,-2){\small RPM-Net}
\end{overpic}
\\[0.05\linewidth]

\begin{overpic}
[width=\textwidth]{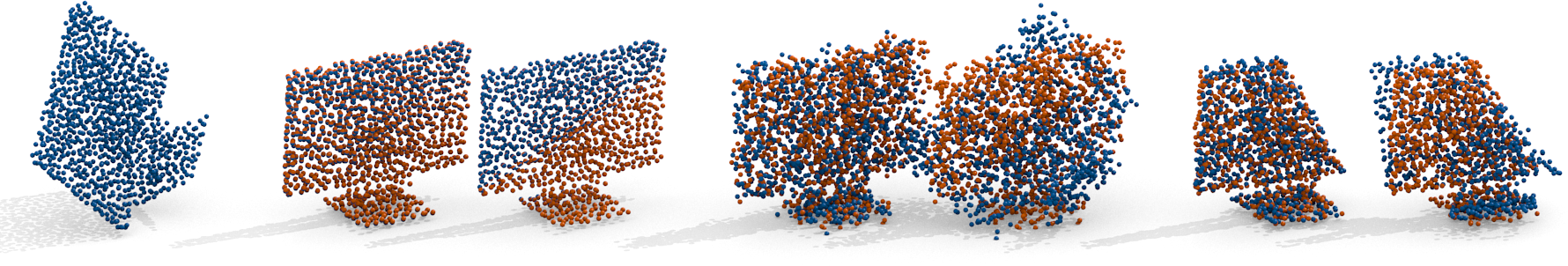}
\put(20,-2){\small Ours}
\put(30,-2){\small RPM-Net}
\put(50,-2){\small Ours}
\put(60,-2){\small RPM-Net}
\put(80,-2){\small Ours}
\put(90,-2){\small RPM-Net}
\end{overpic}
\\[0.05\linewidth]

\end{center}
\caption{\textbf{Point cloud registration.} Additional qualitative examples of RPM-Net \cite{yew2020rpm} and the augmented version with our custom backward pass, see \cref{alg:backward}. The increased stability of the gradients predicted by our algorithm directly translates into more robust generalization results: Both methods are trained and tested on separate subsets of the 40 object categories in ModelNet40 \cite{wu20153d}, see \cite[Section~6]{yew2020rpm} for more details. Both methods yield accurate predictions for the clean test data, as indicated by the corresponding quantitative results in \cref{table:rpm}. On the other hand, our approach shows significant improvements when generalizing to noisy test data and partial inputs.
}
\label{fig:rpm_qual_2}
\end{figure*}

\cref{thm:errorbounds} states that the error of our backward pass in \cref{alg:backward} can be bounded by the error of the forward pass in \cref{eq:forwardpass}. We now assess the magnitude of this error in practice by revisiting the experiments on image barycenter computation and number sorting from \cref{sec:experiments}. Since the ground truth gradients are unknown in general, we define the (approximate) ground truth gradients as $\nabla_\mC\ell^*:=\nabla_\mC\ell^{(\tau_\mathrm{max})},\nabla_\va\ell^*:=\nabla_\va\ell^{(\tau_\mathrm{max})}$ for $\tau_\mathrm{max}:=10,000$. In \cref{fig:gradient_approx}, we compare the error of our approach to AD, averaged over all settings and iterations from the experiments in \cref{fig:img_barycenter} and \cref{fig:number_sorting} respectively. These results show clearly that, on average, our approach produces more accurate gradients than AD. Intuitively, ours yields optimal gradients for a suboptimal forward pass $\mP^*\approx\mS_\lambda^{(\tau)}$, see \cref{thm:errorbounds}, whereas AD yields approximate gradients for an approximate forward pass. To further illustrate this point, we show a qualitative comparison of the gradients of both approaches on the task of image barycenter computation in \cref{fig:gradient_approx_qual}. These results demonstrate the usefulness of our custom backward pass. Our gradients and the ground truth start disagreeing in certain regions for $t\geq50$, but only after the barycenter optimization converges. Compared to the vanilla AD approach, the gradients are much closer to the ground truth and therefore more useful. 

\subsection{MNIST image clustering}\label{subsec:mnistimageclustering}

We mainly consider the barycenter problem described in \cref{subsec:barycenter} a useful toy example to assess the stability of implicit differentiation in comparison to AD. On the other hand, the interpretation as a generic optimization problem allows us to trivially extend it to related tasks like image clustering. To that end, we apply the k-means algorithm which alternates between computing cluster centroids (by minimizing \cref{eq:barycenter}) and assigning all images $\vb_i$ to their closest centroid (defined as the centroid with minimum distance $d$, \cref{eq:barycenterb}). We show results on the 60k images from the MNIST dataset \cite{lecun1998mnist} in \cref{fig:imageclustering}.

\subsection{Details on permutation baselines}\label{subsec:additionalpermutation}

\paragraph{Gumbel-Sinkhorn networks.}
As outlined in \cref{subsec:permutation}, the goal of the Gumbel-Sinkhorn method \cite{mena2018learning} is to learn how to sort a set of input elements $\{x_1,\dots,x_n\}$.
To this end, the cost matrix $\mC$ is parameterized via a permutation-invariant neural network architecture (set encoder), conditioned on the input elements $\{x_1,\dots,x_n\}$. The matrix $\mC$, together with marginals $\va=\vb=\bbone_n$ are then passed through a differentiable Sinkhorn layer.\footnote{Strictly speaking, the choice of marginals $\va=\vb=\bbone_n$ does not fit in our framework, since we require $\va,\vb\in\Delta_n$, see \cref{eq:probsimplex}. However, we can simply use $\va=\vb=\frac{1}{n}\bbone_n$ and scale the resulting transportation plan $\mP$ by a constant factor of $n$.}
The final output $\mP$ is a bistochatic matrix which encodes an approximate permutation. The training objective is a geometric loss, minimizing the distance of the sorted elements $x_i$ to their natural ground-truth ordering. At test time, the Hungarian algorithm \cite{kuhn1955hungarian} is applied to the learned cost matrix $\mC$ to obtain an exact, hard permutation. 

In \cref{subsec:permutation}, we show the concrete application of sorting $n$ real numbers, sampled randomly from the uniform distribution $x_i\sim\mathcal{U}(0,1)$. More specifically, we consider separate training and test sets of $4096$ and $128$ random sequences each and report the error, defined as the proportion of incorrectly placed numbers in the test set, see \cref{fig:number_sorting}. To provide a more complete picture, we report quantitative results on the task of generalizing to different test sets here. Specifically, we train the vanilla GS method and our approach for $\tau=200$ (the maximum number for which GS is below the GPU memory limit) for $n=200$ numbers sampled from $\mathcal{U}(0,1)$. We then investigate the error on test sets sampled from different distributions $x_i\sim\mathcal{U}(s,t)$ with $s<t$ in \cref{table:numbersortinggeneralization}. Even though the variance is quite high, these results indicate that our method evidently helps to reduce overfitting. Note, that the improved generalization is observed for the setting $\tau=200$, $n=200$ where the performance of both methods on the standard test set is almost identical, see \cref{fig:number_sorting}.

\begin{table*}[!h]
\vspace{0.02\linewidth}
\centering
\resizebox{0.8\linewidth}{!}{
\begin{tabular}{l|llll}
& $\mathcal{U}(1,2)$ & $\mathcal{U}(10,11)$ & $\mathcal{U}(100,101)$ & $\mathcal{U}(-1,0)$ \\ \hline
Ours & $0.2964 \pm 0.0744$ & $0.3340 \pm 0.3059$ & $0.3531 \pm 0.1380$ & $0.3552 \pm 0.2116$ \\
Gumbel-Sinkhorn & $0.3163 \pm 0.0976$ & $0.3620 \pm 0.3179$ & $0.3955 \pm 0.1478$ & $0.4678 \pm 0.2526$
\end{tabular}
}
\caption{\textbf{Number sorting generalization.} We assess the capability of our approach (first row) and the vanilla Gumbel-Sinkhorn method (second row) \cite{mena2018learning} to generalize to various test sets $\mathcal{U}(s,t)$ with $s<t$. We train both methods to sort sets of $n=200$ numbers $x_i$ from the uniform distribution on the unit interval $\mathcal{U}(0,1)$ with $\tau=200$ Sinkhorn iterations. For each test set, we show the mean proportion of false predictions, as well as the empirical standard deviation, obtained from $50$ test runs per setting.
}
\label{table:numbersortinggeneralization}
\end{table*}

\paragraph{RPM-Net.}
A number of recent works use the Sinkhorn operator as a differentiable matching module for deep learning \cite{sarlin2020superglue,yew2020rpm,yang2020mapping,liu2020learning,eisenberger2020deep}. The standard strategy of such methods is to use a learnable feature extractor to obtain descriptors $\mF^{X}\in\mathbb{R}^{m\times p},\mF^{Y}\in\mathbb{R}^{n\times p}$ on pairs of input objects $X$ and $Y$ in a siamese manner. We can then define the cost matrix $\mC:=\mD$ as the squared pairwise distance matrix $\mD_{i,j}=\bigl\|\mF_{:,i}^X-\mF_{:,j}^Y\bigr\|_2^2$. For fixed, uniform marginals $\va$ and $\vb$, the Sinkhorn operator then yields a soft matching $\mP=S_\lambda(\mC,\va,\vb)\in\mathbb{R}^{m\times n}$ between the two input objects. The baseline method RPM-Net we consider in \cref{subsec:permutation} uses this methodology to obtain a matching between pairs of input point clouds. As a feature extractor, it uses PointNet \cite{qi2017pointnet} with a custom input task point cloud, see \cite[Sec.~5.1]{yew2020rpm} for more details. Using the obtained soft correspondences, a simple SVD transformation then yields the optimal rigid transformation between the two input point clouds. In order to train their model, RPM-Net uses automatic differentiation of the Sinkhorn layer. We can now demonstrate that replacing AD with our backward algorithm improves the performance. To that end, we revisit the experiments from \cite[Sec.~6]{yew2020rpm}: We use 20 separate object identities of the ModelNet40 dataset \cite{wu20153d} as our train and test set respectively. In \cref{fig:rpm_qual_2}, we now show a qualitative comparison corresponding to the results in \cref{table:rpm} in the main paper. On the standard test set, both approaches produce comparable, high-quality results. On the other hand, our method significantly improves the robustness when generalizing to noisy data or partial views.

\clearpage
\onecolumn

\section{Proofs}
In the following, we provide proofs of Lemma~\ref{thm:kkt}, Lemma~\ref{thm:qpjacobian}, \cref{thm:closedformbackward}, \cref{thm:algorithmequivalence} and \cref{thm:errorbounds} from the main paper.

\subsection{Proof of Lemma~\ref{thm:kkt}}\label{subsec:prooflemmakkt}
\begin{proof}
The function $\mathcal{K}$ contains the KKT conditions corresponding to the optimization problem in \cref{eq:sinkhornoperator}. The proposed identity therefore follows directly from the (strict) convexity of \cref{eq:sinkhornoperator}, see \cite{erlander1990gravity}. Apart from the two equality constraints, \cref{eq:polytopepi} contains additional inequality constraints $P_{i,j}\geq0$. Those are however inactive and can be dropped, because the entropy term in \cref{eq:sinkhornoperator} invariably yields transportation plans in the interior of the positive orthant $P_{i,j}>0$, see \cite[p. 68]{peyre2019computational}.
\end{proof}

\subsection{Proof of Lemma~\ref{thm:qpjacobian}}\label{subsec:prooflemmaqp}
\begin{proof}
The key for proving this statement is applying the implicit function theorem. We start by (a) showing that this indeed yields the identity from \cref{eq:qpjacobian}, and then (b) justify why removing the last $(m+n)$-th equality condition from $\mE$ is necessary.

\begin{enumerate}
    \item[(a)] First of all, we can verify by direct computation that the matrix $\mK$ from \cref{eq:qpjacobian} 
    corresponds to the partial derivatives of the KKT conditions from \cref{eq:kkt}
    \begin{equation}
        \mK=\biggl(\frac{\partial\mathcal{K}(\mCh,\va,\vb,\mPh,\valpha,\vbeta)}{\partial\begin{bmatrix}
            \mPh;\valpha;\vbeta
        \end{bmatrix}}\biggr)_{-l,-l},
    \end{equation}
    where the notation $(\cdot)_{-l,-l}$ means that the last row and column is removed and where $l=mn+m+n$. Furthermore, $\mK$ is invertible (since the solution lies in the interior $\mP_{i,j}>0$, see \cite[p. 68]{peyre2019computational}), and the $m+n-1$ columns of $\mEt$ are linearly dependent, see (b). Consequently, the implicit function theorem states that $\mathcal{K}$ implicitly defines a mapping $(\mCh,\va,\vb)\mapsto(\mPh,\valpha,\vbeta)$ whose Jacobian is
    \begin{equation}
        \mJ=\frac
    {\partial\begin{bmatrix}
        \mPh;\valpha;\vbetat
    \end{bmatrix}}
    {\partial\begin{bmatrix}
        \mCh;-\va;-\vbt
    \end{bmatrix}}=-
    \biggl(\frac{\partial\mathcal{K}}{\partial\begin{bmatrix}
            \mPh;\valpha;\vbeta
        \end{bmatrix}}\biggr)_{-l,-l}^{-1}\underbrace{\biggl(\frac{\partial\mathcal{K}}{\partial\begin{bmatrix}
            \mCh;-\va;-\vb
        \end{bmatrix}}\biggr)_{-l,-l}}_{=\mI_{l-1}}=-\mK^{-1}.
    \end{equation}
    \item[(b)] 
    As part of the proof in (a), we use the fact that the columns of $\mEt$ are linearly independent. Verifying this statement also provides insight as to why removing the  last row of $\vbt,\vbetat$ and the last column of $\mEt$ is necessary. Intuitively, the columns of $\mE$ contain one redundant condition: The identity
    \begin{equation}
        \sum_{i=1}^m\mP_{i,n}=0\iff\mE_{:,m+n}^\top\mPh=0,
    \end{equation}
    follows directly from the other $m+n-1$ conditions. More formally, we can take the kernel
    \begin{equation}\label{eq:redundantcondition}
        \ker(\mE^\top)=\bigl\{\mP\in\mathbb{R}^{m\times n}|\mE^\top\mPh=0\bigr\},
    \end{equation}
    of $\mE^\top\in\mathbb{R}^{m+n\times mn}$ and observe that $\dim\bigl(\ker(\mE^\top)\bigr)=(m-1)(n-1)$, see \cite[Sec.~1]{bolker1972transportation}. The rank–nullity theorem then implies that the dimension of the subspace spanned by the columns of $\mE$ is of dimension $mn-(m-1)(n-1)=m+n-1$. Consequently, removing the redundant condition in \cref{eq:redundantcondition} from $\mE$ yields the reduced $\mEt\in\mathbb{R}^{mn\times m+n-1}$ with $=m+n-1$ linearly independent columns.
\end{enumerate} 
\end{proof}

\subsection{Proof of \cref{thm:closedformbackward}}\label{subsec:closedformbackward}

\begin{proof}
This identity follows trivially from Lemma~\ref{thm:qpjacobian} by applying the chain rule
\begin{equation}
\nabla_{[\mCh;-\va;-\vbt]}\ell=
    \left(\frac
    {\partial\begin{bmatrix}
        \mPh;\valpha;\vbetat
    \end{bmatrix}}
    {\partial\begin{bmatrix}
        \mCh;-\va;-\vbt
    \end{bmatrix}}\right)^\top
    \nabla_{[\mPh;\valpha;\vbetat]}\ell=-\mK^{-1}\nabla_{[\mPh;\valpha;\vbetat]}\ell=\mK^{-1}\begin{bmatrix}
        -\nabla_{\mPh}\ell\\\mathbf{0}
    \end{bmatrix}.
\end{equation}
The last equality holds, since the loss $\ell$ does not depend on the dual variables $\valpha$ and $\vbeta$, see \cref{fig:overview}. 
\end{proof}

\subsection{Proof of \cref{thm:algorithmequivalence}}\label{subsec:algorithmequivalence}

\begin{proof}
We want to show that the gradients $\nabla_\mC\ell,\nabla_\va\ell,\nabla_\vb\ell$ obtained with \cref{alg:backward} are equivalent to the solution of \cref{eq:closedformbackward}. To that end, we start by applying the Schur complement trick to the block matrix $\mK$. This yields the expression
\begin{equation}
    \bigl(\mK^{-1}\bigr)_{:,1:mn}=\begin{bmatrix}
        \lambda^{-1}\diagP\bigl(\mI_{mn}+\mEt\bigl(\mEt^\top\diagP\mEt\bigr)^{-1}\mEt^\top\diagP\bigr)\\
        \bigl(\mEt^\top\diagP\mEt\bigr)^{-1}\mEt^\top\diagP
    \end{bmatrix},
\end{equation}
for the first $mn$ columns of its inverse.
In the next step, we can insert this expression in \cref{eq:closedformbackward} and invert the linear system of equations
\begin{equation}
    \begin{bmatrix}\nabla_{\mCh}\ell\\-\nabla_{[\va;\vbt]}\ell\end{bmatrix}=\mK^{-1}\begin{bmatrix}-\nabla_{\mPh}\ell\\\mathbf{0}\end{bmatrix}=-\bigl(\mK^{-1}\bigr)_{:,1:mn}\nabla_{\mPh}\ell.
\end{equation}
Further simplification yields the following identities for the gradients of $\mC,\va$ and $\vb$
\begin{align}\label{eq:schurgrad}
    \begin{bmatrix}\nabla_{\mCh}\ell\\\nabla_{[\va;\vbt]}\ell\end{bmatrix}=&\begin{bmatrix}
        -\lambda^{-1}\diagP\bigl(\mI_{mn}-\mEt\bigl(\mEt^\top\diagP\mEt\bigr)^{-1}\mEt^\top\diagP\bigr)\nabla_{\mPh}\ell\\
        \bigl(\mEt^\top\diagP\mEt\bigr)^{-1}\mEt^\top\diagP\nabla_{\mPh}\ell
    \end{bmatrix}\nonumber\\=&\begin{bmatrix}
        -\lambda^{-1}\bigl(\diagP\nabla_{\mPh}\ell-\diagP\mEt\nabla_{[\va;\vbt]}\ell\bigr)\\
        \bigl(\mEt^\top\diagP\mEt\bigr)^{-1}\mEt^\top\diagP\nabla_{\mPh}\ell
    \end{bmatrix},
\end{align}
where the latter equality results from substituting the obtained expression for $\nabla_{[\va;\vbt]}\ell$ in the first block row.
In the remainder of this proof, we can show line by line that these expressions yield \cref{alg:backward}. The main idea is to first compute the second block row identity in \cref{eq:schurgrad} and then use the result to eventually obtain $\nabla_{\mCh}\ell$ from the first block row:
\begin{enumerate}
    \item[\cref{alg:lnT}] The first line defines the matrix $\mT:=\mP\odot\nabla_\mP\ell$ via the Hadamard product $\odot$. In  vectorized form it corresponds to the expression
    \begin{equation}\label{eq:Th}
        \mTh=\diagP\nabla_{\mPh}\ell.
    \end{equation}
    \item[\cref{alg:lnTt}] As detailed in Lemma~\ref{thm:qpjacobian}, we remove the last equality condition from $\mE$ to obtain $\mEt$. Equivalent considerations require us to introduce the truncated versions $\mTt,\mPt$ of $\mT,\mP$. 
    \item[\cref{alg:vtavttb}] The operator $\mE^\top$ then maps $\mTh$ to the vector
    \begin{equation}\label{eq:Ttorowcolsum}
        \mE^\top\mTh=
        \begin{bmatrix}\bbone_n\otimes\mI_m&\mI_n\otimes\bbone_m\end{bmatrix}^\top\mTh=
        \begin{bmatrix}(\bbone_n^\top\otimes\mI_m)\mTh\\(\mI_n\otimes\bbone_m^\top)\mTh\end{bmatrix}=
        \begin{bmatrix}\mT\bbone_n\\\mT^\top\bbone_m\end{bmatrix},
    \end{equation}
    that contains its row and column sums. In terms of the truncated $\mEt$, the last row of \cref{eq:Ttorowcolsum} gets removed, thus
    \begin{equation}\label{eq:Ttorowcolsumreduced}
    \mEt^\top\mTh=
        \begin{bmatrix}\mT\bbone_n\\\mTt^\top\bbone_m\end{bmatrix}=
        \begin{bmatrix}\vta\\\vttb\end{bmatrix}.
    \end{equation}
    \item[\cref{alg:gradagradb}] A direction computation reveals that
    \begin{align}\label{eq:epematrix}
        \mE^\top\diagP\mE=&
        \begin{bmatrix}
        (\bbone_n^\top\otimes\mI_m)\diagP(\bbone_n\otimes\mI_m) &
        (\bbone_n^\top\otimes\mI_m)\diagP(\mI_n\otimes\bbone_m) \\
        (\mI_n\otimes\bbone_m^\top)\diagP(\bbone_n\otimes\mI_m) & 
        (\mI_n\otimes\bbone_m^\top)\diagP(\mI_n\otimes\bbone_m)\end{bmatrix}\nonumber\\=&
        \begin{bmatrix}\diag(\mP\bbone_n) & \mP \\ \mP^\top & \diag(\mP^\top\bbone_m) \end{bmatrix}=
        \begin{bmatrix}\diag(\va) & \mP \\ \mP^\top & \diag(\vb) \end{bmatrix}.
    \end{align}
    The linear system in \cref{alg:gradagradb} of \cref{alg:backward} therefore yields the gradients $\nabla_{[\va;\vbt]}\ell$ by inserting \cref{eq:Th}, \cref{eq:Ttorowcolsumreduced} and (the reduced version of) \cref{eq:epematrix} into the second block row identity from \cref{eq:schurgrad}.
    \item[\cref{alg:gradbresidual}] We can expand $\nabla_{\vbt}\ell$ to $\nabla_{\vb}\ell$ by appending zero $\nabla_{b_n}\ell=0$ as the last entry. Since $\vb$ is constrained to the probability simplex $\Delta_n$ this gradient is exact for all entries, see the discussion in \cref{subsec:practicalconsiderations}.
    \item[\cref{alg:Umatrix}] Having computed the gradients $\nabla_{[\va;\vb]}\ell$, we can now insert them in the first row of \cref{eq:schurgrad}. Here, the reduced and the original expressions are equivalent
    \begin{equation}\label{eq:mEt}
        \mEt\nabla_{[\va;\vbt]}\ell=\mE\nabla_{[\va;\vb]}\ell
    \end{equation}
    because \cref{alg:gradbresidual} specifies $\nabla_{b_n}\ell=0$. Thus,
    \begin{equation}
        \mEt\nabla_{[\va;\vbt]}\ell=\begin{bmatrix}\bbone_n\otimes\mI_m&\mI_n\otimes\bbone_m\end{bmatrix}\nabla_{[\va;\vb]}\ell=
    \bbone_n\otimes\nabla_{\va}\ell+\nabla_{\vb}\ell\otimes\bbone_m=:\mUh,
    \end{equation}
    defines the vectorized version of $\mU$ from \cref{alg:Umatrix}.
    \item[\cref{alg:gradm}] Putting everything together, we insert the identities from \cref{eq:Th} and \cref{eq:mEt} into the first block row of \cref{eq:schurgrad}
    \begin{equation}
        \nabla_{\mCh}\ell=
        -\lambda^{-1}\bigl(\diagP\nabla_{\mPh}\ell-\diagP\mEt\nabla_{[\va;\vbt]}\ell\bigr)=
        -\lambda^{-1}\bigl(\mTh-\diagP\mUh\bigr),
    \end{equation}
    which is equivalent to the matrix-valued expression in \cref{alg:gradm}.
\end{enumerate}
\end{proof}

\subsection{Proof of \cref{thm:errorbounds}}\label{subsec:errorboundsproof}

\begin{proof}
The key for constructing the error bounds in \cref{eq:errorboundsa} and \cref{eq:errorboundsb} is finding a bound for the first-order derivatives $\frac{\partial\nabla_{\mCh}\ell}{\partial\mPh}$ and $\frac{\partial\nabla_{[\va;\vb]}\ell}{\partial\mPh}$. For brevity, we introduce the short-hand notation $\mPb:=\diagP$. Furthermore, we define the projection of $\vx$ onto the column space of $\mEt$ as
\begin{equation}\label{eq:projectionoperator}
\PiE\vx:=
\underset{\vy\in\mathrm{span}(\mEt)}{\arg\min}\|\vx-\vy\|^2_{\mPb}=
\mEt\underset{\vz\in\mathbb{R}^{m+n-1}}{\arg\min}\|\vx-\mEt\vz\|^2_{\mPb},
\end{equation}
where $\langle\cdot,\cdot\rangle_{\mPb}:=\langle\mPb^{\frac{1}{2}}\cdot,\mPb^{\frac{1}{2}}\cdot\rangle_{2}$. In matrix notation, \cref{eq:projectionoperator} reads
\begin{equation}
\PiE=\mEt\bigl(\mEt^\top\mPb\mEt\bigr)^{-1}\mEt^\top\mPb.
\end{equation}
Using \cref{eq:gradientla} and \cref{eq:gradientlb} from the proof of \cref{thm:algorithmequivalence}, we can then rewrite the backward pass compactly as
\begin{subequations}
\begin{equation}\label{eq:gradientlacompact}
    \nabla_{[\va;\vbt]}\ell=\mEt^\dagger\PiE\nabla_{\mPh}\ell,\quad\text{and}
\end{equation}
\begin{equation}\label{eq:gradientlbcompact}
    \nabla_{\mCh}\ell=
        -\lambda^{-1}\bigl(\mPb(\mI-\PiE)\dPl\bigr),
\end{equation}
\end{subequations}
The first identity follows from $\mEt^\dagger\mEt=\mI$, since the columns of $\mEt$ are linearly independent (see part (b) of the proof of Lemma~\ref{thm:qpjacobian} in \cref{subsec:prooflemmaqp}). Direct substitution of \cref{eq:gradientla} into \cref{eq:gradientlb} immediately yields \cref{eq:gradientlbcompact}. To differentiate $\nabla_{[\va;\vbt]}\ell$ and $\nabla_{\mCh}\ell$, we apply the chain rule which in turn requires a closed-form expression for the derivative of the projection operator $\PiE$. Since it is defined as the solution of an optimization problem, we apply the implicit function theorem to the gradient of the objective in \cref{eq:projectionoperator}, i.e.
\begin{equation}
    \nabla_{\vz}\biggl(\frac{1}{2}\|\vx-\mEt\vz\|^2_{\mPb}\biggr)=\mEt^\top\mPb\mEt\vz-\mEt^\top\mPb\vx=\mathbf{0}.
\end{equation}
The Jacobian of the mapping $\mPh\mapsto\PiE\vx$ can therefore be written in terms of the IFT as
\begin{equation}
    \frac{\partial\PiE\vx}{\partial\mPh}=\mEt(\mEt^\top\mPb\mEt)^{-1}\mEt^\top\diag\bigl(\vx-\PiE\vx\bigr),
\end{equation}
This auxiliary result implies that the Jacobians of the mappings $\mPh\mapsto\nabla_{[\va;\vbt]}\ell$ and $\mPh\mapsto\nabla_{\mCh}\ell$ defined in \cref{eq:gradientlacompact} and \cref{eq:gradientlbcompact} are
\begin{subequations}
\begin{align}
    \begin{split}
    \frac{\partial\nabla_{[\va;\vbt]}\ell}{\partial\mPh}=&
    \mEt^\dagger\frac{\partial}{\partial\mPh}\PiE\dPl=(\mEt^\top\mPb\mEt)^{-1}\mEt^\top\diag\bigl((\mI-\PiE)\dPl\bigr)+\mEt^\dagger\PiE\ddPl,\quad\text{and}
\end{split}\\
\begin{split}
    \frac{\partial\nabla_{\mCh}\ell}{\partial\mPh}=&
    -\lambda^{-1}\biggl(\diag\bigl((\mI-\PiE)\dPl\bigr)-\PiE^\top\diag\bigl((\mI-\PiE)\dPl\bigr)+\mPb(\mI-\PiE)\ddPl\biggr)\\=&
    -\lambda^{-1}\biggl(\bigl(\mI-\PiE^\top\bigr)\diag\bigl((\mI-\PiE)\dPl\bigr)+\mPb(\mI-\PiE)\ddPl\biggr).
\end{split}
\end{align}
\end{subequations}

In order to bound the errors of these two gradients, we first derive an upper bound for the norm of the operator $\PiE$. An important insight is that we can precondition $\PiE$ via $\mPb^{\frac{1}{2}}$
\begin{equation}
\mPb^{\frac{1}{2}}\PiE\mPb^{-\frac{1}{2}}\vx=
\underset{\vy\in\mathrm{span}(\mPb^{\frac{1}{2}}\mEt)}{\arg\min}\|\vx-\vy\|^2_2,
\end{equation}
which results in an orthogonal projection $\mPb^{\frac{1}{2}}\PiE\mPb^{-\frac{1}{2}}$. Since such projections have a spectral radius of at most $1$, we can bound the norm of $\PiE$ as
\begin{equation}
    \bigl\|\PiE\bigr\|_2\leq
    \bigl\|\mPb^{-\frac{1}{2}}\bigr\|_2\bigl\|\mPb^{\frac{1}{2}}\PiE\mPb^{-\frac{1}{2}}\bigr\|_2\bigl\|\mPb^{\frac{1}{2}}\bigr\|_2\leq
    \bigl\|\mPb^{-\frac{1}{2}}\bigr\|_2\bigl\|\mPb^{\frac{1}{2}}\bigr\|_2,
\end{equation}
and equivalently show for the complementary projector $\mI-\PiE$ that 
\begin{equation} 
\bigl\|\mI-\PiE\bigr\|_2\leq
    \bigl\|\mPb^{-\frac{1}{2}}\bigr\|_2\bigl\|\mPb^{\frac{1}{2}}(\mI-\PiE)\mPb^{-\frac{1}{2}}\bigr\|_2\bigl\|\mPb^{\frac{1}{2}}\bigr\|_2\leq
    \bigl\|\mPb^{-\frac{1}{2}}\bigr\|_2\bigl\|\mPb^{\frac{1}{2}}\bigr\|_2.
\end{equation}
The Jacobians of the backward pass can then be bounded as
\begin{subequations}
\begin{align}
    \begin{split}
    \biggl\|\frac{\partial\nabla_{[\va;\vbt]}\ell}{\partial\mPh}\biggr\|_F\leq&
    \bigl\|(\mEt^\top\mPb\mEt)^{-1}\mEt^\top\bigr\|_2\bigl\|(\mI-\PiE)\dPl\bigr\|_2+\bigl\|\mEt^\dagger\PiE\ddPl\bigr\|_F\\=&
    \bigl\|\mEt^\dagger\PiE\mPb^{-1}\bigr\|_2\bigl\|(\mI-\PiE)\dPl\bigr\|_2+\bigl\|\mEt^\dagger\PiE\ddPl\bigr\|_F\\\leq&
    \bigl\|\mEt^\dagger\bigr\|_2\bigl\|\mPb^{-\frac{1}{2}}\bigr\|_2^2\bigl\|\mI-\PiE\bigr\|_2\bigl\|\dPl\bigr\|_2+\bigl\|\mEt^\dagger\bigr\|_2\bigl\|\PiE\bigr\|_2\bigl\|\ddPl\bigr\|_F\\\leq&
    \bigl\|\mEt^\dagger\bigr\|_2\bigl\|\mPb^{-\frac{1}{2}}\bigr\|_2\bigl\|\mPb^{\frac{1}{2}}\bigr\|_2\biggl(\bigl\|\mPb^{-\frac{1}{2}}\bigr\|_2^2\|\dPl\bigr\|_2+\bigl\|\ddPl\bigr\|_F\biggr)\\\leq&
    \kappa\sqrt{\frac{\sigma_+}{\sigma_-}}\biggl(\frac{1}{\sigma_-}C_1+C_2\biggr),\quad\text{and}
\end{split}\\
\begin{split}
    \biggl\|\frac{\partial\nabla_{\mCh}\ell}{\partial\mPh}\biggr\|_F\leq&
    \lambda^{-1}\bigl\|\mI-\PiE^\top\bigr\|_2\bigl\|\mI-\PiE\bigr\|_2\bigl\|\dPl\bigr\|_2+ \lambda^{-1}\bigl\|\mPb(\mI-\PiE)\bigr\|_2\bigl\|\ddPl\bigr\|_F\\\leq&
    \lambda^{-1}\bigl\|\mPb^{-\frac{1}{2}}\bigr\|_2^2\bigl\|\mPb^{\frac{1}{2}}\bigr\|_2^2\bigl\|\dPl\bigr\|_2+ \lambda^{-1}\bigl\|\mPb^{\frac{1}{2}}\bigr\|_2^2\bigl\|\ddPl\bigr\|_F\\\leq&
    \lambda^{-1}\sigma_+\biggl(\frac{1}{\sigma_-}C_1+C_2\biggr),
\end{split}
\end{align}
\end{subequations}
where the constants $\sigma_-,\sigma_+,C_1,C_2>0$ are as defined in \cref{thm:errorbounds}, and where we use the identity
\begin{equation}
    \bigl\|\mA\diag(\vb)\bigr\|_F\leq
    \bigl\|\mA\bigr\|_2\bigl\|\diag(\vb)\bigr\|_F=
    \bigl\|\mA\bigr\|_2\bigl\|\vb\bigr\|_2.
\end{equation}
As a direct consequence, we obtain the bounds from \cref{eq:errorboundsa} and \cref{eq:errorboundsb}, since the bounded derivatives imply the Lipschitz continuity of the differentiable map $\mPh\mapsto\nabla_{[\mCh;\va;\vb]}\ell$.

\end{proof}

\end{document}